\documentclass[pdflatex,sn-mathphys-num]{sn-jnl}

\usepackage{graphicx}%
\usepackage{multirow}%
\usepackage{amsmath,amssymb,amsfonts}%
\usepackage{amsthm}%
\usepackage{mathrsfs}%
\usepackage[title]{appendix}%
\usepackage{xcolor}%
\usepackage{textcomp}%
\usepackage{manyfoot}%
\usepackage{booktabs}%
\usepackage{algorithm}%
\usepackage{algorithmicx}%
\usepackage{algpseudocode}%
\usepackage{listings}%
\usepackage{upgreek}%
\usepackage{mathtools} %
\usepackage{caption}
\usepackage{subcaption}
\usepackage{placeins}
\theoremstyle{thmstyleone}%
\theoremstyle{thmstyletwo}%

\theoremstyle{thmstylethree}%

\begin{document}

\title[Article Title]{Causal Intelligence for Constraint‑Aware Intervention Design to Induce State Transitions}


\author[1]{\fnm{Zixuan} \sur{Song}}\email{zixuan.song@merck.com}
\equalcont{These authors contributed equally to this work.}

\author[1]{\fnm{Uwe} \sur{Mueller}}\email{uwe.mueller@merck.com}

\author*[1]{\fnm{Dimitris V.} \sur{Manatakis}}\email{dimitrios.manatakis@merck.com}
\equalcont{These authors contributed equally to this work.}

\affil[1]{\orgname{MRL, Merck \& Co., Inc.}, \orgaddress{\street{33 Avenue Louis Pasteur}, \city{Boston}, \postcode{02115}, \state{MA}, \country{USA}}}


\abstract{
Driving a system from one state to another through targeted interventions is a fundamental challenge in science, yet most predictive models offer limited mechanistic insight and no principled framework for decision-making. Here we present \textbf{COAST} (\textbf{C}ausally \textbf{O}ptimal \textbf{A}ctions for \textbf{S}tate \textbf{T}ransitions), a causal‑intelligence approach for the \emph{in‑silico} design of constrained interventions that induce user‑defined state transitions. Given data characterizing source and target states, COAST learns context‑specific causal graphs and structural causal models, attributes observed distributional shifts to mechanism‑level causal drivers, and introduces a novel constraint‑aware multi‑objective optimization formulation that balances transition efficacy, intervention complexity, and target‑state stability. The approach is modular and domain‑agnostic, integrating feature selection, causal discovery, causal modeling, and intervention identification and evaluation through interchangeable components. Across synthetic benchmarks and real biological datasets, COAST recovers key causal drivers and identifies robust single‑ and multi‑target intervention strategies that achieve desired state transitions, accompanied by transparent mechanistic rationales to guide experimental validation.
}




\maketitle

\section{Introduction}\label{sec1}

Identifying minimal yet effective interventions to drive a complex system from one state to another is a fundamental challenge across science and engineering \cite{Scheffer2009}. In biomedicine, state transitions encompass progression from health to disease, shifts across disease stages or tumor subtypes, and lineage decisions during differentiation, phenomena commonly represented as networks of interacting entities \cite{Correia2025,Yang2020,Panditrao2022}. These dynamics are governed by causal principles. Learning the causal relationships that structure such mechanisms is therefore a fundamental prerequisite for models that not only explain observations but also reliably predict the effects of single or combinatorial interventions. This capability enables \emph{in-silico} identification and prioritization of interventions that achieve desired outcomes, such as halting pathological progression or reversing disease states, while reducing experimental burden and accelerating discovery.

Conventional AI/ML systems are optimized to capture statistical associations and maximize predictive accuracy. By design, they do not identify causes and therefore rarely provide the mechanistic insight required for hypothesis-driven reasoning or counterfactual \emph{what-if} analysis \cite{Pearl2009,Scholkopf2021}. Robust decision support in complex systems requires methods that embed causal reasoning at their core, moving beyond pattern recognition toward explicit models of mechanisms, interventions, and their consequences.

A central challenge in biology and drug discovery is identifying the minimal set of targets, often in combination, that causally drive a specific biological process or disease mechanism. Addressing this challenge requires models that can (i) distinguish causal drivers from correlated effects, (ii) attribute observed state differences to specific mechanisms, and (iii) reason explicitly about the outcomes of constrained interventions. This defines an \emph{inverse design} problem, fundamentally distinct from forward response prediction \cite{Lotfollahi2019,lotfollahi2023predicting,Hetzel2022chemCPA,roohani2024predicting,Bunne2023}. Rather than predicting how a system responds to a known perturbation, the goal is to identify which interventions will drive it toward a desired state while respecting mechanistic and practical constraints.

Recent computational frameworks have made substantial progress in predicting transcriptional responses to genetic and chemical perturbations. Methods such as scGen \cite{Lotfollahi2019}, CPA \cite{lotfollahi2023predicting}, GEARS \cite{roohani2024predicting}, and CellOT \cite{Bunne2023} employ deep generative or optimal-transport formulations to extrapolate perturbation effects and demonstrate generalization to unseen conditions. These approaches are highly effective for response prediction and data augmentation, but are not designed to address the inverse design problem. They neither infer experiment-specific causal structure nor provide mechanism-level attribution that supports interpretable causal decision-making.

PDGrapher targets the inverse design objective more directly by formulating the identification of effective perturbations as a supervised learning task that leverages large-scale genetic and chemical perturbation libraries and graph-based architectures to predict perturbagens capable of inducing a desired transcriptional state change \cite{Gonzalez2025}. This paradigm achieves strong empirical performance when paired perturbation–response data at scale are available. However, PDGrapher does not perform causal structure learning, instead operating on predefined biological interaction networks and learning predictive relationships between perturbations and transcriptional outcomes rather than explicitly estimating causal effects or constructing a mechanistic graph. Consequently, it does not identify which upstream regulators causally explain the observed differences between source and target states or attribute state transitions to specific dysregulated mechanisms. As a result, it does not support experiment-specific mechanistic reasoning, such as \emph{why a given state difference exists} and \emph{which causal pathways drive it} that is required to prioritize targets with biological rationale. Furthermore, practical constraints on interventions, such as druggability, toxicity thresholds, or combinatorial feasibility, are not explicitly incorporated into its objective. The questions of \emph{which mechanisms causally explain the transition between two states}, and \emph{how to optimally intervene under explicit biological and practical constraints}, therefore remain open.

We introduce \textbf{COAST} (\textbf{C}ausally \textbf{O}ptimal \textbf{A}ctions for \textbf{S}tate \textbf{T}ransitions), a modular, domain‑agnostic causal‑intelligence framework for identifying constraint‑aware interventions that induce user‑defined state transitions. Given data characterizing source and target states, COAST is designed to identify single or combinatorial interventions whose \emph{causal effects} are sufficient to induce a desired state transition, while explicitly taking into account mechanistic, biological, and practical feasibility constraints.

To enable this optimization, COAST (i) performs context-specific feature selection to define a biologically meaningful modeling universe from the data at hand; (ii) learns context-specific causal graph(s) and associated structural causal models that capture the mechanisms operative in the system under study; and (iii) attributes observed distributional shifts between source and target states to mechanism-level causal drivers. These components jointly support the core innovation of COAST: a multi-objective causal optimization formulation that directly searches over feasible intervention sets under explicit constraints, balancing transition efficacy, target-state stability, and intervention sparsity. Importantly, feature selection, causal discovery, and modeling components are modular and interchangeable, enabling COAST to accommodate diverse data modalities and modeling assumptions without altering the underlying optimization problem.

Although COAST is domain-agnostic by construction, we focus here on biomedical applications, where identifying causal drivers and interpretable intervention strategies is central to hypothesis generation and experimental prioritization. In this setting, COAST enables principled intervention design by linking mechanism-level attribution to constrained optimization, allowing trade-offs between transition efficacy and intervention complexity while ensuring that proposed interventions satisfy biological and practical feasibility constraints. The framework produces inspectable mechanistic rationales that directly inform experimental prioritization. By solving this optimization across a range of regularization strengths ($\lambda$), COAST yields a spectrum of intervention strategies that vary in density; targets that recur consistently across sparse and dense solutions provide an intrinsic robustness signal whose persistence is independent of any particular regularization choice. Together, these properties enable COAST to address a class of problems not targeted by large‑scale perturbation models: deriving principled, mechanistically grounded intervention strategies directly from observational or interventional data, without reliance on perturbation atlases. This makes COAST particularly well suited to settings where such libraries are unavailable, but data characterizing source and target states can be obtained, including early‑stage disease studies and under-characterized biological contexts, where observational data can be collected but systematic perturbation resources and mechanistic knowledge remain limited.

\section{Results}\label{sec2}
We evaluate COAST by assessing its ability to identify correct and biologically meaningful interventions that drive systems between distinct states. First, using controlled synthetic benchmarks with known ground truth, we test whether COAST recovers the true causal drivers and intervention targets underlying the observed state shift. Second, using Perturb‑seq data, we evaluate whether COAST correctly identifies the actual genetic perturbations responsible for the observed transcriptional changes. Finally, using single‑cell RNA‑seq data capturing transitions between cell states, we assess whether COAST proposes interventions whose downstream effects activate the biological pathways governing the corresponding state transition.

\subsection{COAST overview}

\begin{figure}[!tbp]
  \centering
  \includegraphics[width=\linewidth]{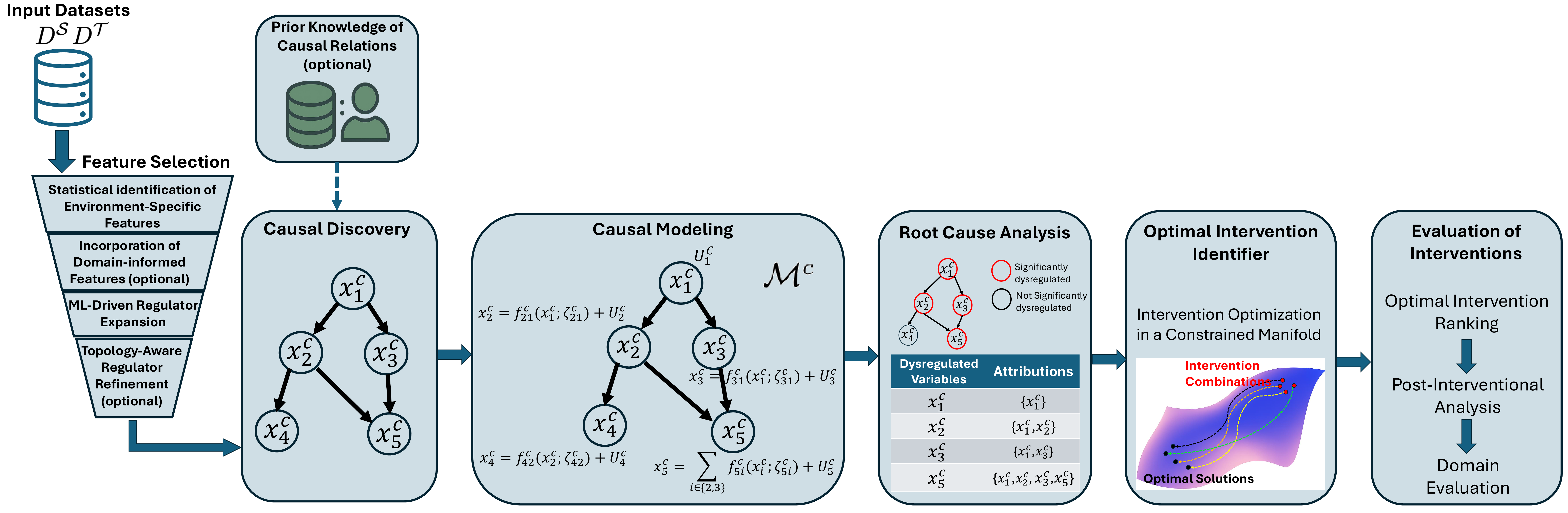}
  \caption{\textbf{Overview of the COAST framework.} COAST identifies single or combinatorial interventions that drive a system from a source state $\mathcal{S}$ to a target state $\mathcal{T}$ through six tightly coupled modules: feature selection, causal discovery, causal modeling, root cause analysis, optimal intervention identifier, and evaluation of interventions. Starting from source and target data ($D^{\mathcal{S}}$ and $D^{\mathcal{T}}$), COAST first performs context‑specific feature selection and learns causal graph structure(s). Given the inferred graph structure(s) and the corresponding data, COAST then fits context‑specific causal models $\mathcal{M}^{c}$, $c \in \lbrace \mathcal{S}, \mathcal{T} \rbrace$. COAST subsequently attributes observed state differences to mechanism‑level causal drivers and solves a novel constraint‑aware multi‑objective optimization problem to propose feasible intervention sets. Candidate interventions are then ranked, simulated \emph{in-silico}, and evaluated using domain‑specific criteria.}
  \label{flowdiagram}
\end{figure}

COAST is a causal‑intelligence approach for identifying single or combinatorial interventions that drive a system from a source state $\mathcal{S}$ to a desired target state $\mathcal{T}$ (Fig.~\ref{flowdiagram}). It implements a principled end‑to‑end causal pipeline composed of six tightly coupled modules, each producing outputs that serve as necessary inputs to subsequent stages. Together, these modules form a causally coherent chain that maps observational or interventional data to actionable intervention strategies. Full methodological details are provided in Section~\ref{sec3}.

The pipeline begins with \textbf{feature selection}, which defines a context-specific modeling universe by identifying features that exhibit significant differential behavior between $\mathcal{S}$ and $\mathcal{T}$, augmented optionally with prior-knowledge variables and upstream regulators inferred through regulatory network analysis. This focused feature set is then passed to \textbf{causal discovery}, which learns the directed acyclic graph(s) (DAGs) encoding putative cause-effect relationships among the selected features. Rather than pooling $D^{\mathcal{S}}$ and $D^{\mathcal{T}}$, COAST treats them as distinct regimes, supporting multi-environment structure learning strategies that can recover either a shared causal backbone or state-specific graphs depending on domain assumptions. The inferred graph(s) is subsequently used by \textbf{causal modeling} to fit state-specific structural causal models (SCMs) $\mathcal{M}^{\mathcal{S}}$ and $\mathcal{M}^{\mathcal{T}}$, which parameterize the causal mechanisms operative in each state and serve as \emph{in-silico} surrogates for simulating the downstream effects of candidate interventions.

Building on the learned causal structure, \textbf{root cause analysis} employs mechanism-level attribution to quantify how changes in each variable's conditional distribution contribute to the observed state difference, yielding a ranked list of causal drivers that defines and narrows the intervention search space. These ranked drivers directly inform the \textbf{optimal intervention identifier}, which formulates a novel constrained multi-objective optimization problem that outputs minimal intervention sets whose post-interventional distribution maximally aligns $\mathcal{S}$ with $\mathcal{T}$, jointly trading off transition efficacy, intervention sparsity, and target-state stability, while enforcing practical feasibility constraints such as druggability and relative-change bounds. By solving this optimization across a range of regularization strengths ($\lambda$), COAST produces a spectrum of solutions, from which intervention strategies that recur consistently across regularization values are identified as robust. Finally, \textbf{evaluation of interventions} ranks candidate strategies by transition percentage, quantifies downstream effects via \emph{in-silico} simulation under the fitted SCMs, and contextualizes results through domain-specific criteria --- in biomedical settings, pathway enrichment and disease-association analyses --- to prioritize interventions for experimental follow-up.

\begin{figure}[!tbp]
  \centering
  \includegraphics[width=\linewidth]{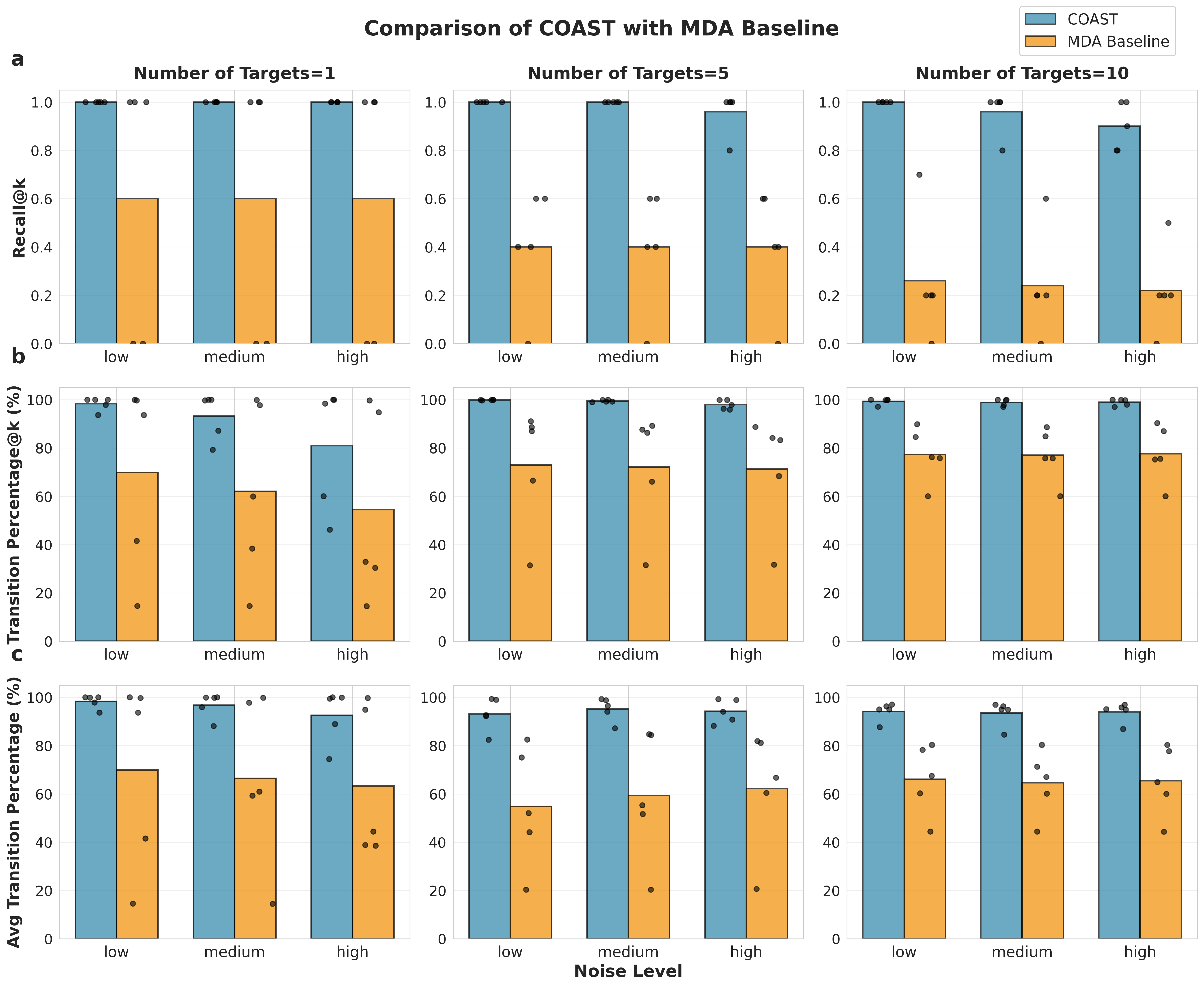}
  
  \caption{\textbf{Results on synthetic datasets.} Comparison of COAST and the marginal distributional analysis (MDA) baseline on 100-node graphs across noise levels (low: $\sigma=1$, medium: $\sigma=3$, high: $\sigma=5$). Within each row, panels correspond to different numbers of ground-truth intervention targets ($k \in \{1, 5, 10\}$). Bar heights denote means over five random seeds, and black dots show individual seed values. \textbf{(a)} Recall@$k$, measuring the fraction of selected targets that are true intervention targets, where selection is based on the top-$k$ ranked targets (with $k$ matched to the ground-truth number of targets). COAST achieves near-perfect recall across settings, whereas MDA exhibits substantially lower recall, particularly as $k$ increases. \textbf{(b)} Transition percentage@$k$, quantifying the fraction of the source-to-target distributional shift explained by the identified intervention, evaluated at the solution whose intervention size matches the ground truth. COAST consistently outperforms MDA by 21--31\%. \textbf{(c)} Average transition percentage, computed over all non-zero solution sizes along the regularization path (different $\lambda$ values), providing a comprehensive assessment across different intervention budgets. COAST maintains stable performance above 92\% regardless of noise level or number of targets, while MDA shows greater variability and systematically lower values.}
  
  \label{synthetic_plots}
\end{figure}

\subsection{Validation on controlled synthetic benchmarks}

We evaluate COAST on synthetic datasets and compare it to a non-causal baseline based on \emph{marginal distributional analysis} (MDA). MDA scores each variable by its univariate source--target shift (e.g., absolute mean difference or an associated test statistic) and selects intervention targets as the top $k$ variables with the largest MDA scores, with $k$ matched to the ground-truth number of intervened variables.

For each experimental configuration, we simulate a single structural causal model (SCM) with a known directed acyclic graph (DAG) and fixed causal mechanisms, which are shared between the source and target regimes. We generate source state observational samples by drawing from the SCM of the observational (non-interventional) distribution. To generate the target state, we randomly select $k$ intervention targets and apply ground-truth (surgical) interventions to these variables; the resulting perturbations propagate to downstream nodes according to the same SCM, thereby inducing a shifted target distribution. The procedures for data generation and intervention follow \cite{Zhu2020Causal}, with standardization performed as in \cite{zhang2023active}. We then evaluate the performance of each method (COAST and MDA) by assessing whether it (i) recovers the true intervention targets and (ii) explains the induced transition from the source to the target distribution. We vary the number of intervention targets ($k \in \{1,5,10\}$) and the noise level (low, medium, high; $\sigma \in \{1,3,5\}$), repeating each configuration across five random seeds. Fig. \ref{synthetic_plots} reports results for 100-node graphs; additional results for graph sizes 10 and 500 are provided in the supplementary material (see Fig.~\ref{fig:synthetic_results_10} and Fig.~\ref{fig:synthetic_results_500}).

\paragraph{Evaluation metrics}
We report three complementary metrics. \textbf{Recall@$k$} is the fraction of selected targets that are true intervention targets, with $k$ fixed to the ground-truth intervention size. For COAST, targets are ranked by a three-stage procedure: (i) frequency of appearance across the regularization path (across different $\lambda$ values), (ii) attribution score of distributional change, and (iii) random tie-breaking. For MDA, variables are ranked by their univariate marginal shift scores and the top-$k$ are selected. \textbf{Transition percentage@$k$} measures the fraction of the total source--target distributional shift explained by the identified intervention, evaluated at the solution whose intervention size matches the ground truth (see Section~\ref{OpIntIdent}). \textbf{Average transition percentage} averages transition percentage over all non-zero solution sizes along the regularization path, thereby capturing performance across a range of intervention budgets (see Appendix \ref{secA1}).

Figure~\ref{synthetic_plots}a summarizes recall@$k$ across noise levels and intervention sizes. COAST achieves near-perfect recall in most settings, with mean recall@$k$ of 1.00 for $k=1$ across all noise levels, 0.99 for $k=5$ (decreasing modestly to 0.96 at high noise), and 0.95 for $k=10$. In contrast, MDA yields substantially lower recall, averaging 0.6 for $k=1$, 0.4 for $k=5$, and 0.24 for $k=10$. Across all configurations of different numbers of ground truth intervention targets and different noise levels, COAST improves recall over MDA by 40--74\%, highlighting the benefit of causal structure and mechanism-level attribution over purely marginal criteria for identifying intervention targets.

Figure~\ref{synthetic_plots}b reports transition percentage@$k$ at the solution whose intervention size matches the ground truth. COAST achieves consistently high transition percentages@$k$, ranging from 81\% in the single-target, high-noise regime to above 98\% in all multi-target settings ($k \in \{5,10\}$). MDA attains transition percentages@$k$ between 55\% and 78\%, trailing COAST by 21--31\%. The gap is most pronounced in the single-target regime: at medium noise, COAST achieves a mean transition percentage@$k$ of 93\%, compared to 62\% for MDA. Note the reason why COAST obtains a lower transition percentage@$k$ in single-target cases is the state transition is intrinsically difficult when there is only one variable to intervene, especially when there is large noise; on the other hand, with multiple variables to intervene, there will be more flexibility as they may compensate for the noise.

Figure~\ref{synthetic_plots}c reports average transition percentage across the regularization path. COAST attains mean values between 93\% and 98\% across all configurations, whereas MDA ranges from 55\% to 70\%, with improvements in the range of 28--38\% separately. Similar results are observed for 10-node and 500-node graphs (see supplementary material). Together, these indicate that causal-model-guided optimization yields more accurate and more robust intervention identification than baselines based solely on marginal distributional shifts.

\subsection{Recovery of true perturbation targets on Perturb--seq data}

To evaluate COAST on real biological data, we applied it to a Perturb-CITE-seq dataset from \cite{frangieh2021multimodal}, which profiled gene expression in a melanoma cell line under pooled CRISPR perturbations across three experimental conditions (control, IFN-$\gamma$, and co-culture). Similar to \cite{zhang2023active}, we used only the control screen and focused on a curated set of 36 genes selected based on cofunctional modules and coregulated transcriptional programs identified in the original study \cite{frangieh2021multimodal}, which are involved in cancer immune evasion pathways including interferon-gamma signaling, antigen presentation, and cell cycle regulation. As multi-target interventional samples are extremely scarce in this dataset, with most such interventions having no more than one sample, we selected single-gene knockouts with the largest sample sizes for validation.

Following \cite{zhang2023active}, we used the greedy sparsest permutation (GSP) algorithm with partial correlation tests and learnt a causal graph over the 36 genes (Fig. \ref{dag36}). A structural causal model (SCM) was then fitted on the observational (unperturbed) data from 5,039 control cells with no detected guide RNAs, serving as the source state model for all subsequent analyses.

We demonstrate COAST's ability to identify ground truth perturbation targets through three case studies: knockout of CTSD (Cathepsin D), knockout of TGFB1 (Transforming Growth Factor Beta 1), and knockout of B2M (Beta-2-Microglobulin). Results are shown in Table \ref{tab1} and \ref{tab2}.

\begin{table}[!tbp]
  \centering
  \caption{Single-gene targets identified by COAST as the most frequently selected along the regularization path in three Perturb-seq case studies.}
  \label{tab1}
  \begin{tabular}{|l|l|c|c|c|}
    \hline
    Study  & Single-target ($x_i$) & Recall@1 & Transition percentage (\%) & Intervention value \\ \hline
    Case 1 & CTSD               & 1                    & 74.8                      & $-2.65\,\downarrow$ \\ \hline
    Case 2 & TGFB1              & 1                    & 57.0                      & $-1.87\,\downarrow$ \\ \hline
    Case 3 & B2M                & 1                    & 27.7                      & $-0.92\,\downarrow$ \\ \hline
  \end{tabular}
\end{table}

\begin{table}[!tbp]
  \centering
  \caption{Multi-gene intervention sets identified by COAST yielding the highest transition percentages in three Perturb-seq case studies.}
  \label{tab2}
  \begin{tabular}{|l|l|c|c|l|}
    \hline
    Study & Intervention Set ($\mathcal{I}$) & Transition percentage (\%) & Intervention value(s) \\ \hline
    Case 1 & (CTSD, DNMT1, NPC2) & 82.0 & $(-2.82\,\downarrow,\ 0.34\,\uparrow,\ 0.42\,\uparrow)$ \\ \hline
    Case 2 & (TGFB1, CDK6) & 65.0 & $(-1.91\,\downarrow,\ 0.55\,\uparrow)$ \\ \hline
    Case 3 & (B2M, CTSB, CDK6, NPC2) & 65.7 &
    $(-1.49\,\downarrow,\ 0.27\,\uparrow,\ 0.49\,\uparrow,\ 0.91\,\uparrow)$ \\ \hline
  \end{tabular}
\end{table}

In all three case studies, COAST successfully identified the true knockout gene as the top-ranked root cause and the optimal single-node intervention target (Table \ref{tab1}). In case 1, the source state consisted of 5,039 unperturbed control cells and the target state consisted of 170 cells with a single-gene CTSD knockout. COAST assigned the highest distribution-change attribution to CTSD, selected it as the most stable intervention target with a transition percentage of 75\% (i.e., intervening on CTSD alone explains roughly 75\% of the shift), and returned an optimal intervention value of $-2.65(\downarrow)$, indicating strong downregulation consistent with a knockout; CTSD remained selected along the regularization path as $\lambda$ increased, underscoring its dominance. In case 2, where the target state consisted of 164 cells with a single-gene TGFB1 knockout, COAST likewise ranked TGFB1 highest, found it to dominate across all sparsity levels, and reported a transition percentage of 57\% with an optimal intervention of $-1.87(\downarrow)$. In case 3, the target state consisted of 132 cells with a single-gene B2M knockout, and COAST assigned B2M the highest attribution score (1.58), well separated from the next best genes CTSB (0.062) and CDK6 (0.061); regularization confirmed B2M as the dominant target (it was the last gene regularized to zero across $\lambda$ values), and the optimal interventions were consistently downregulations ($\downarrow$).

On top of ranking individual genes and identifying the optimal single-gene intervention, COAST can also propose multi-gene intervention sets that yield higher transition percentages. As Table \ref{tab2} shows, in case 1, a CTSD intervention combined with two additional genes achieves a transition percentage above 80\%, and case 2 shows a similar pattern, that a two-gene intervention that includes the ground-truth knockout TGFB1 increases the transition percentage by about 8\% relative to the best single-gene intervention. In case 3, the optimal multi-gene intervention identified by COAST, which consists of 4 intervened genes, substantially improved the transition percentage compared with intervening on B2M alone. These improvements may reflect common off-target effects in CRISPR experiments and the ability of multi-target interventions to average out noises in the data. Together, these results demonstrate COAST's advantage in identifying combinations of interventions that achieve superior transitions compared with single-gene interventions.

Overall, across three single-gene knockout case studies, COAST correctly identified the true causal genes and returned coherent single-gene interventions, while additionally identifying multi-gene intervention sets that substantially improve transition percentages in some cases. These results show that COAST can recover known perturbation targets from real Perturb-seq data using only an observational causal model and the observed distributional shift, and that its ability to propose multi-target interventions helps capture more distributed or noisy responses, yielding more effective transitions than single-gene interventions alone.

\subsection{Biologically coherent intervention design for cell-state transitions}

We applied COAST to a real single-cell RNA-Seq dataset from \cite{chen2017}, who profiled the adult mouse hypothalamus and identified 45 cell types through clustering analysis. We focused on 5,282 cells belonging to two non-neuronal clusters: oligodendrocyte precursor cells (OPCs) and myelinating oligodendrocytes (MOs), which represent two distinct stages of oligodendrocyte maturation. Following standard single-cell RNA-Seq preprocessing, genes expressed in fewer than 30\% of cells were filtered out, leaving 1,046 genes for analysis. Raw gene expression was transformed using $\log(\text{TPM} + 1)$.

\begin{figure}[!tbp]
  \centering
  \includegraphics[width=\linewidth]{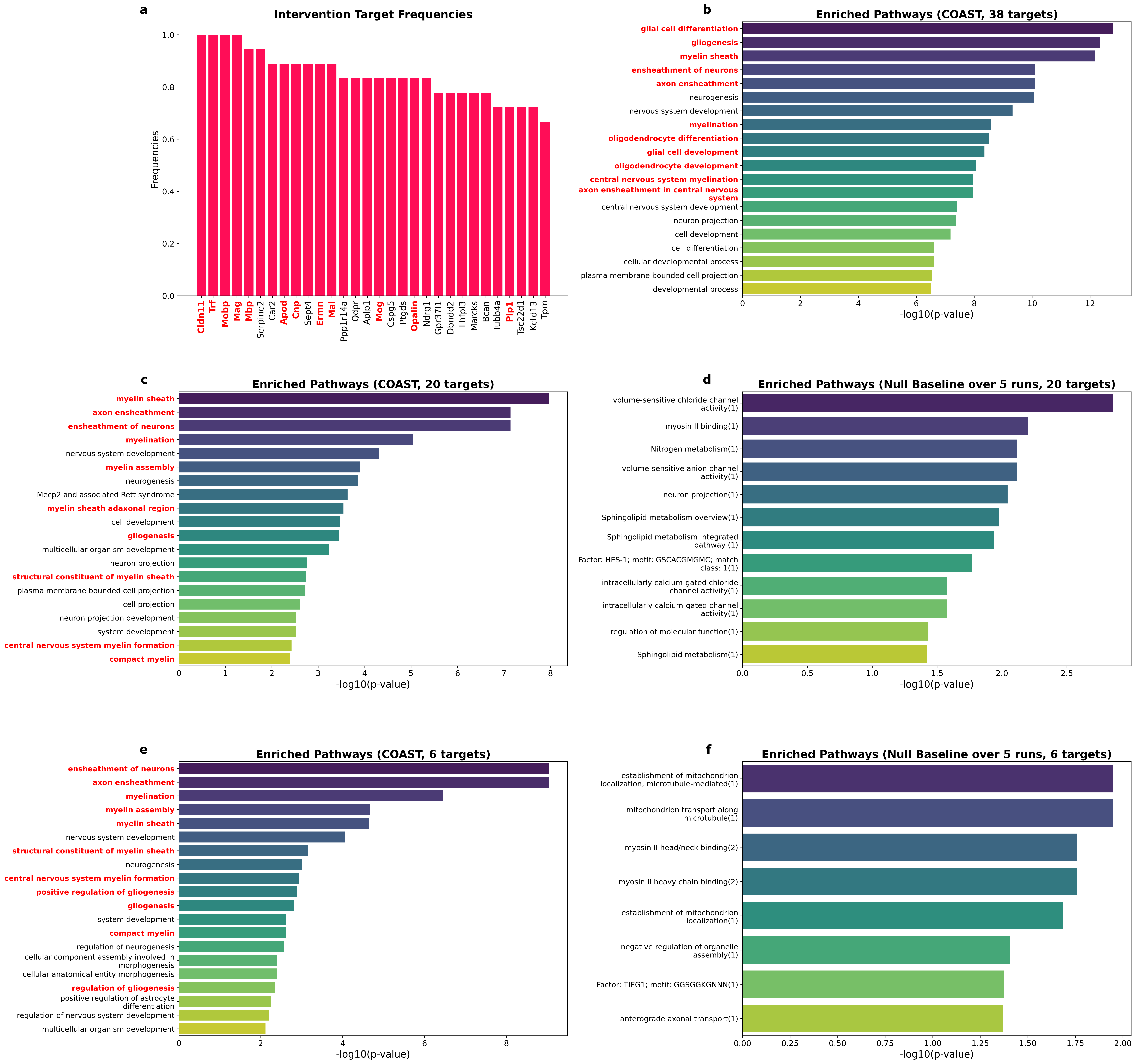}
  \caption{\textbf{COAST intervention targets and pathway enrichment compared to null intervention baselines in the OPC-to-MO transition.}
(a) Frequencies of intervention target genes across the regularization path. Frequencies were computed as the fraction of unique solutions in which each gene retained a non-zero intervention value. Genes are ranked by decreasing frequency, and the top 30 genes are shown. Known oligodendrocyte maturation and myelination markers are bolded and colored in red. (b) Top 20 enriched pathways (g:Profiler) among downstream genes significantly affected by in silico perturbation of all 38 COAST-identified intervention targets. Bar lengths represent $-\log_{10}(p\text{-value})$. Pathways related to oligodendrocyte maturation are bolded and colored in red. (c,d) Pathway enrichment comparison between COAST-identified intervention targets (c) and null intervention targets (d) for interventions comprising 20 genes. Null intervention results were obtained by aggregating g:Profiler enrichment across five independent random samplings using Fisher’s method; the number appended to each GO term indicates the number of runs in which the term was detected. (e,f) Corresponding comparison for interventions comprising 6 genes, illustrating the persistence of pathway specificity under increasingly constrained intervention budgets.}
  \label{chen_plots}
\end{figure}

\textbf{Feature selection:} Differential expression (DE) analysis was performed using Welch's $t$-test between the OPC (source) and MO (target) populations. Applying thresholds of FDR-adjusted $p$-value $< 0.05$ and $|\text{log fold-change}| > 2$ yielded 262 target DE genes. After the third step of the feature selection pipeline (see Section \ref{Feature Selection}), which augmented the DE gene set with inferred upstream regulators and applied network-based prioritization, 263 genes were retained for downstream causal analysis.

\textbf{Causal discovery and modeling:} The IMaGES (Independent Multi-sample Greedy Equivalence Search) algorithm was employed to learn a single causal DAG from the combined OPC and MO datasets \cite{Ramsey2010fMRI}. Two separate structural causal models (SCMs) were then fitted on the source-state (OPC) and target-state (MO) data using the inferred DAG, with additive noise models assigned to each node. These SCMs serve as digital-twin surrogates for subsequent \emph{in-silico} experimentation.

\textbf{Root cause analysis and optimal intervention identification:} Mechanism-level attribution scores were computed via Shapley value-based decomposition (see Section \ref{RCA}), and through adaptive selection the top 38 genes were identified as candidate intervention targets. Using the regularization-based optimization approach, COAST generated a series of intervention solutions along the regularization path, trading off transition effectiveness against sparsity (see supplementary material Fig. \ref{results_reg_chen}).

Analysis of gene frequencies across the regularization path revealed a stable core of intervention targets (Fig. \ref{chen_plots}a). To compute these frequencies, we counted the fraction of unique solutions in which each gene retained a non-zero intervention value across the full range of regularization penalties. Four genes---\textit{Cldn11}, \textit{Trf} (transferrin), \textit{Mobp}, and \textit{Mag}---appeared in 100\% of unique solutions, identifying them as the most robust intervention targets. These were followed by \textit{Mbp} and \textit{Serpine2} at 94.4\% frequency, \textit{Car2} (carbonic anhydrase 2), \textit{Apod} (apolipoprotein D), \textit{Cnp}, \textit{Sept4}, \textit{Ermn}, and \textit{Mal} at 88.9\% frequency, and a group of 8 genes including \textit{Mog}, \textit{Opalin}, \textit{Ptgds}, and \textit{Ndrg1} at 83.3\% frequency. Critically, the highest-frequency genes are dominated by known oligodendrocyte maturation and myelination markers. \textit{Mobp} distinguishes myelinating oligodendrocytes from precursor stages \citep{chen2017, marques2016}. \textit{Cldn11} is a tight junction protein essential for CNS myelin formation \citep{morita1999}. \textit{Trf} is required for iron delivery to oligodendrocytes during myelination \citep{ortiz2004, connor1996}, while \textit{Apod} is involved in protection from oxidative stress through the control of lipid peroxidation \citep{ganfornina2008}. \textit{Mag}, \textit{Mal}, \textit{Mog}, \textit{Mbp}, \textit{Plp1}, and \textit{Cnp} are canonical myelin structural proteins and myelination regulators \citep{baumann2001, nave2014}. \textit{Ermn} \citep{brockschnieder2006} and \textit{Opalin} \citep{golan2008} are markers specific to mature myelinating oligodendrocytes. The consistent selection of these biologically validated genes across different levels of sparsity provides strong evidence that COAST identifies interventions grounded in the true underlying biology of the OPC-to-MO transition.

\textbf{Evaluation of interventions:} To assess whether the identified interventions induce biologically coherent downstream effects, we performed \emph{in-silico} perturbation experiments followed by pathway enrichment analysis using g:Profiler \cite{Kolberg2023gProfiler}.

With the full set of 38 intervention genes (see also supplementary material Fig.~\ref{results_reg_chen}), the \emph{in-silico} perturbation produced significant expression changes in 53 downstream genes. Pathway enrichment analysis of these affected genes revealed terms that are highly specific to the OPC-to-MO transition (Fig. \ref{chen_plots}b): glial cell differentiation ($p = 1.67 \times 10^{-13}$), gliogenesis ($p = 4.42 \times 10^{-13}$), myelin sheath ($p = 6.77 \times 10^{-13}$), ensheathment of neurons ($p = 7.77 \times 10^{-11}$), myelination ($p = 2.73 \times 10^{-9}$), oligodendrocyte differentiation ($p = 3.13 \times 10^{-9}$), and structural constituent of myelin sheath ($p = 1.49 \times 10^{-6}$) etc. These terms directly correspond to the hallmark biological processes of oligodendrocyte maturation---the progressive acquisition of a myelinating phenotype involving myelin membrane assembly and axon ensheathment \citep{emery2015}---providing strong evidence that the COAST-identified interventions faithfully recapitulate the transcriptional program driving the OPC-to-MO transition.

To assess robustness under more practical intervention budgets, we repeated the in-silico perturbation with smaller intervention sets identified by COAST, and the core myelination-related pathways remained significantly enriched. With 20 intervention genes (Fig. \ref{chen_plots}c), the top enriched terms included myelin sheath ($p = 1.07 \times 10^{-8}$), axon ensheathment ($p = 7.23 \times 10^{-8}$), myelination ($p = 9.25 \times 10^{-6}$), and oligodendrocyte differentiation ($p = 2.05 \times 10^{-2}$). With only 6 intervention genes (Fig. \ref{chen_plots}e), the enrichment remained striking: ensheathment of neurons ($p = 9.06 \times 10^{-10}$), myelination ($p = 3.47 \times 10^{-7}$), myelin assembly ($p = 2.14 \times 10^{-5}$), myelin sheath ($p = 2.24 \times 10^{-5}$), and central nervous system myelination ($p = 1.78 \times 10^{-2}$) etc. These results demonstrate that even when the intervention set is substantially reduced, with correspondingly lower transition percentages, the downstream effects remain centered on the biological processes most specific to the OPC-to-MO differentiation, indicating that COAST prioritizes the most mechanistically critical targets first.

\paragraph{Comparison with null intervention targets}
To verify that the observed pathway activation and transition efficiencies arise from alignment with the inferred causal structure rather than from arbitrary gene perturbations, we compared COAST-identified targets against null intervention targets of matched cardinality. Null intervention targets were generated by uniform sampling over the admissible intervention space and serve as a statistical null model in which perturbations are, in expectation, not aligned with the causal pathways governing the OPC-to-MO transition. For practical intervention budgets (20 and 6 genes), five independent null intervention target sets were generated using distinct random seeds and evaluated using the same in-silico intervention and downstream pathway enrichment pipeline.

To obtain robust enrichment estimates across the five null intervention replicates, g:Profiler results were combined by taking the union of all enriched GO terms across runs. For terms identified in multiple runs, $p$-values were aggregated using Fisher’s method, where the test statistic $\chi^2 = -2\sum_{i=1}^{k} \ln(p_i)$ was evaluated against a $\chi^2$ distribution with $2k$ degrees of freedom, and $k$ denotes the number of runs in which the term was detected.

COAST-identified targets consistently achieved substantially higher transition percentages than matched null intervention baselines, with even more pronounced differences in downstream pathway activation. For interventions comprising 20 genes, COAST yielded 44 enriched pathways dominated by myelination-related terms (including myelin sheath, axon ensheathment, and myelination, as described above), whereas the Fisher-combined results across five null-intervention target sets produced only 12 enriched pathways—none associated with myelination or oligodendrocyte biology (Fig.~\ref{chen_plots}d); the most significant combined terms reflected generic cellular functions, including volume-sensitive chloride channel activity ($p = 1.40 \times 10^{-3}$), myosin II binding ($p = 6.29 \times 10^{-3}$), nitrogen metabolism ($p = 7.64 \times 10^{-3}$), and regulation of molecular function ($p = 3.69 \times 10^{-2}$).

A similarly pronounced disparity was observed for smaller interventions of 6 genes. COAST identified 33 enriched pathways, again including highly significant myelination-related processes (such as ensheathment of neurons and myelination), whereas aggregated results from five null-intervention target sets yielded only 8 enriched pathways (Fig.~\ref{chen_plots}f), led by mitochondrion transport along microtubule ($p = 1.13 \times 10^{-2}$) and myosin II heavy chain binding ($p = 1.74 \times 10^{-2}$), which are unrelated to oligodendrocyte differentiation. Notably, even after aggregating results across 5 independent null interventions, no myelination- or oligodendrocyte-related pathway emerged.

Together, these results demonstrate that COAST’s performance is driven by its identification of causally effective intervention targets within the inferred DAG, rather than by perturbations that do not align with the causal pathways underlying the OPC-to-MO transition.

\section{Discussion}\label{sec4}

Designing interventions that reliably drive complex systems between distinct states is a central challenge across scientific domains, particularly in settings where mechanistic understanding, interpretability, and feasibility constraints are critical. In this work, we introduced COAST, a causal‑intelligence framework that, given data characterizing source and target states, learns context‑specific causal graph(s) and structural causal models, attributes observed distributional shifts to mechanism‑level causal drivers, and formulates intervention design as a constraint‑aware multi‑objective optimization problem balancing transition efficacy, intervention complexity, and target‑state stability. Unlike predictive or response‑forecasting models, COAST addresses an inverse causal design problem: determining \emph{which} actions to take, and \emph{how} to take them, in order to causally drive a system toward a desired target state.

Across controlled synthetic benchmarks and multiple real biological datasets, COAST consistently identified interventions that were not only effective in inducing state transitions, but also mechanistically grounded and robust to modeling and regularization choices. On synthetic data with known ground truth, COAST reliably recovered the true intervention targets and achieved near‑complete explanation of the induced distributional shift, substantially outperforming marginal, non‑causal baselines -- particularly as noise levels increased, a regime that closely mirrors the variability encountered in real biological systems. On Perturb‑seq datasets, COAST accurately recovered the true genetic perturbations by leveraging inferred causal models and observed state differences, without access to perturbation labels. In single-cell RNA-seq data capturing oligodendrocyte maturation, COAST identified intervention targets whose downstream effects activated pathway programs directly tied to the biological processes governing the transition, substantially outperforming null intervention baselines, with these effects persisting even as intervention budgets were reduced.

A central distinguishing feature of COAST is its principled integration of feature selection, causal discovery, causal modeling, root cause analysis, and  constrained optimization within a unified, end‑to‑end framework. Rather than treating intervention design as a purely predictive or heuristic exercise, COAST leverages learned causal structure and models to attribute observed state differences to underlying mechanisms and to reason explicitly about how interventions propagate through the system. By incorporating persistence along the regularization path as an intrinsic robustness criterion, COAST prioritizes intervention strategies not only by transition effectiveness but also by their persistence across intervention sparsity levels, reducing sensitivity to arbitrary tuning choices. This shifts the focus from identifying a single nominally “optimal” solution to characterizing a set of plausible, mechanistically grounded intervention strategies that explicitly balance effectiveness, sparsity, and robustness.

COAST’s formulation emphasizes practical relevance by incorporating multiple feasibility constraints directly into the optimization objective, which enable biologically realistic intervention proposals, such as limiting perturbations to actionable variables, bounding relative changes to preserve essential system functions, and maintaining target‑state stability. In biomedical contexts, this design is particularly important for minimizing unintended effects and ensuring that computational recommendations remain compatible with experimental and therapeutic constraints. This consideration is critical for target identification and biomarker discovery, where neglecting practicability can yield findings that are statistically significant yet biologically implausible or difficult to reproduce experimentally.

At the same time, COAST is intentionally modular and domain‑agnostic. Individual components for feature selection, causal discovery, mechanism attribution, and optimization can be replaced without altering the overall framework, allowing the method to adapt to different data modalities, modeling assumptions, and scientific questions. This flexibility contrasts with approaches that rely on large perturbation atlases or fixed predictive architectures: COAST can operate in data‑sparse settings where observational measurements of source and target states are available, but systematic perturbation data are not.

A key limitation of COAST is that the quality of its recommendations depends on the fidelity of the learned causal models. Although COAST supports multi‑environment causal discovery and incorporates graph falsification procedures, reliable causal structure learning from finite observational or interventional data remains challenging, particularly in high‑dimensional regimes. Future extensions could improve robustness by explicitly modeling uncertainty over causal graphs or by more formally integrating partial mechanistic priors and domain knowledge into the discovery and inference process.

More broadly, COAST is not intended to replace predictive perturbation models, but rather to complement them. While predictive and generative approaches excel at forecasting system responses for prespecified candidate interventions, COAST addresses the upstream problem of intervention selection and mechanistic prioritization. Integrating these paradigms, for example, by using causally guided intervention design to propose candidate perturbations that are subsequently evaluated using high‑fidelity predictive models, represents a promising direction for future research.

In summary, COAST demonstrates that causal modeling and constrained optimization can be combined to address the inverse problem of intervention design in complex systems. By grounding intervention strategies in learned causal mechanisms, enforcing domain‑relevant constraints, and emphasizing robustness across solution paths, COAST provides a principled foundation for mechanistically informed intervention design. We expect this causal‑intelligence perspective to be valuable not only in biomedicine, but also in other domains where state transitions must be induced reliably under uncertainty and practical constraints.

\section{Methods}\label{sec3}

In this section we present the technical details of each module of the COAST framework. Throughout, we use \textit{italic} lowercase (uppercase) letters for vectors (matrices), regular font for scalars, and calligraphic letters for directed acyclic graph (DAG), functions, sets, and states.

COAST aims to identify single or combinatorial intervention targets, together with their optimal magnitudes, that induce a transition from a source state $\mathcal{S}$ to a target state $\mathcal{T}$. To this end, COAST integrates a sequence of modular components—\emph{Feature Selection}, \emph{Causal Discovery}, \emph{Causal Modeling}, \emph{Root Cause Analysis}, an \emph{Optimal Intervention Identifier}, and \emph{Evaluation of Interventions}—which together enable principled intervention design. Each module is described in detail below.

\subsection{Feature Selection}\label{Feature Selection}
This module receives as input the source and target datasets, $D^{\mathcal{S}} \in \mathbb{R}^{\mathrm{n_\mathcal{S}} \times \mathrm{p}}$ and $D^{\mathcal{T}} \in \mathbb{R}^{\mathrm{n_\mathcal{T}} \times \mathrm{p}}$ (where $\mathrm{n_\mathcal{S}}$, $\mathrm{n_\mathcal{T}}$ are the number of samples, and $\mathrm{p}$ is the number of features\footnote{Throughout the manuscript we use \emph{features} and \emph{variables} interchangeably.}; $[p]\coloneqq\{1,\dots,\mathrm{p}\}$ denotes the index set of all features), and selects the features to be used downstream for causal discovery and causal modeling. Restricting attention to a set of relevant and informative features reduces computational complexity by shrinking the search space and mitigates estimation noise by excluding weak or spurious variables. This, in turn, improves statistical efficiency and the stability of learned causal graph structures, while enhancing the transparency and interpretability of the resulting causal model \cite{GuyonElisseeff2003,FanLv2010,MeinshausenBuhlmann2006,Peters2017}.
The feature selection workflow comprises four sequential steps (see Fig. \ref{flowdiagram}).

\textbf{Step 1: Statistical identification of environment‑specific features.}
We first identify variables that exhibit statistically significant differences between the source and target datasets, $D^{\mathcal{S}}$ and $D^{\mathcal{T}}$. To this end, univariate hypothesis tests are applied to each variable: Welch’s $t$‑test is used when approximate normality is satisfied, and the Mann–Whitney $U$ test otherwise \cite{Sheskin2011}. For categorical variables, appropriate alternatives such as the $\chi^2$ test or Fisher’s exact test are employed \cite{Sheskin2011}. To account for multiple hypothesis testing, $p$‑values are adjusted using procedures such as the Benjamini–Hochberg method to control the false discovery rate (FDR) \cite{Sheskin2011}. In high‑dimensional \emph{omics} settings (e.g., transcriptomics, proteomics, metabolomics), we instead rely on established differential‑analysis frameworks, including \texttt{DESeq2}, \texttt{edgeR}, and \texttt{limma‑voom}, which explicitly model count‑based noise and mean-variance relationships \cite{DESeq2,edgeR,limmavoom}.

\textbf{Step 2 (optional): Incorporation of domain‑informed features.}
The statistically identified variables are optionally augmented with features informed by prior domain knowledge, e.g. curated genes or molecular markers of known biological relevance. This step enables the inclusion of variables that may not exhibit strong marginal effects yet are hypothesized to play a mechanistic role. The union of statistically selected and domain‑informed variables defines the subset of ``key'' features $[t]\subset[p]$.

\textbf{Step 3: ML-driven regulator expansion.} Identify additional features (putative regulators) that may causally influence on the key feature set $[t]$. 

\textit{Biomedical instantiation:} COAST employs \textsc{GENIE3} to infer regulatory dependencies by learning an ensemble of tree‑based models and deriving a weighted adjacency matrix \cite{GENIE3}. For each ``key'' feature $\mathrm{t}\in[t]$, the top $\mathrm{N}$ candidate regulators are selected according to their inferred importance scores. The union of the ``key'' features and their corresponding top‑ranked regulators defines the expanded candidate feature set $[r]$, which is passed to the subsequent (optional) step.  

\textit{Framework generality:} this module is fully modular. \textsc{GENIE3} can be replaced with alternative feature‑selection or regulator‑inference procedures, such as elastic‑net or stability‑selection neighborhoods, mutual‑information based screening etc. \cite{ZouHastie2005,MeinshausenBuhlmann2010, CoverThomas2006}, that can provide a ranked list of candidate regulators for each ``key'' feature $\mathrm{t}\in[t]$.

\textbf{Step 4 (optional): Topology-aware regulator refinement.} For \emph{omics} inputs, COAST optionally refines the set of top‑ranked regulators identified in Step~3 by incorporating network topology, thereby further reducing the regulator set carried forward for causal modeling. Starting from the candidate features identified in the previous steps, we re‑estimate regulatory associations using \textsc{GENIE3} and compute, for each regulator-target pair $\mathrm{(r,t)}$ with $\mathrm{t}$ drawn from the key features identified in Step 2, a hybrid prioritization score $\mathrm{h_{rt}}$:

\begin{equation}
\mathrm{h_{rt}} = \alpha\,\mathrm{w_{rt}} + (1-\alpha)\,\mathrm{b_{r}},
\label{feature_score}
\end{equation}
\\
where $\mathrm{w_{rt}}$ denotes the normalized \textsc{GENIE3} edge weight capturing the predictive influence of regulator $\mathrm{r}$ on key feature $\mathrm{t}$, and $\mathrm{b_r}$ denotes the betweenness centrality of regulator $\mathrm{r}$ in the inferred regulatory network.

For each key feature $\mathrm{t}$, candidate regulators are ranked according to $\mathrm{h_{rt}}$, and a user‑specified number $\mathrm{K}$ of top‑scoring regulators (e.g., the top 3 per feature) is retained. This per‑feature ranking yields a refined subset of influential regulators that balances predictive strength and topological importance. Betweenness centrality quantifies the extent to which a regulator lies on shortest paths connecting otherwise weakly coupled network modules, thereby identifying \emph{bottleneck} regulators that disproportionately control information flow. In biological regulatory networks, such bottlenecks are often more indicative of functional importance than highly connected hubs, particularly in directed settings where causal flow is well defined \cite{Freeman1977,Nithya2023,Yu2007Bottlenecks}. From an intervention standpoint, targeting high‑betweenness regulators enables coordinated modulation of multiple downstream pathways, allowing system‑level state transitions to be achieved with fewer, more focused interventions \cite{Nithya2023,Yu2007Bottlenecks}.

The coefficient $\alpha$ controls the relative contribution of predictive and topological signals and is specified \emph{a priori} to reflect domain preferences rather than learned from data. Unless otherwise stated, we use an equal weighting of the two signals.

The output of this step is a consolidated candidate feature set $[q]$, defined as the union of the key features $[t]$ and the subset of top regulators retained per key feature. This reduced and topology‑informed feature set is subsequently used for causal structure learning and intervention design.

\subsection{Causal Discovery}\label{Causal Discovery}
Causal discovery aims to infer cause-effect relations among variables from observational data and, when available, interventional data. We represent causal structure as a directed acyclic graph (DAG) $\mathcal{G}=(\mathcal{V},\mathcal{E})$, where $\mathcal{V}$ denotes variables and $\mathcal{E}\subseteq\mathcal{V}\times\mathcal{V}$ directed edges encode direct causal influence. Under the causal Markov condition and faithfulness, conditional independence patterns constrain admissible structures and enable systematic structure learning \cite{Pearl2009,Spirtes2000}.

A range of structure learning algorithms can be used to estimate $\mathcal{G}$ \cite{Glymour2019}, including constraint-based methods (e.g., PC, FCI) that leverage conditional independence tests \cite{Spirtes2000}, score-based methods (e.g., GES, NOTEARS) that optimize penalized likelihood or information criteria under acyclicity constraints \cite{Chickering2003,Zheng2018}, and functional causal model approaches (e.g., LiNGAM) that impose additional assumptions to improve identifiability \cite{Shimizu2006}. With purely observational data, the learned structure is generally identifiable only up to a Markov Equivalence Class (MEC). Incorporating interventional or multi-context information can refine identifiability and orient additional edges \cite{GIES}. In practice, background knowledge (required/forbidden edges) can be incorporated to constrain the search space and improve recovery accuracy \cite{Tetrad}.

COAST focuses on settings with two environments, corresponding to a \emph{source} state $\mathcal{S}$ and a \emph{target} state $\mathcal{T}$ (e.g., disease and healthy). Rather than naïvely pooling samples from $\mathcal{S}$ and $\mathcal{T}$ and learning a single-regime graph, which can induce statistical dependencies that do not hold within either state, COAST supports multi-environment structure learning strategies that explicitly leverage both datasets.\\
\\
Based on domain assumptions, COAST supports two complementary regimes:

\begin{enumerate}
\item \textbf{Shared-backbone discovery (single graph).}
When $\mathcal{S}$ and $\mathcal{T}$ are assumed to reflect the \emph{same underlying regulatory system} (e.g., the same cell type), COAST adopts a shared-backbone assumption: the \emph{interaction topology is stable} across states (invariant parent sets), while \emph{regulatory strengths and noise} may differ between $\mathcal{S}$ and $\mathcal{T}$. This regime naturally assumes $\mathcal{T}$ as a \emph{perturbation effect} (soft intervention) of the $\mathcal{S}$ that alters causal mechanisms without rewiring the graph, consistent with invariance-based perspectives on multi-environment causality. Accordingly, COAST learns a single causal backbone $\mathcal{G}$ using multi-sample extensions of greedy equivalence search such as IMaGES scoring across datasets \cite{Ramsey2010fMRI}, and then fits two state-specific SCM parameterizations ($\mathcal{M}^{\mathcal{S}}$ and $\mathcal{M}^{\mathcal{T}}$) on the same backbone (one for $\mathcal{S}$ and one for $\mathcal{T}$) to capture state-dependent mechanism changes.\\

\item \textbf{State-specific discovery (two graphs).}
When genuine \emph{rewiring} between $\mathcal{S}$ and $\mathcal{T}$ is plausible (e.g., cancer lineage switching), COAST can learn two related graphs $\mathcal{G}^{\mathcal{S}}$ and $\mathcal{G}^{\mathcal{T}}$. In this regime, COAST supports separate or joint estimation approaches (e.g. jointGES) that share statistical strength across states while allowing a number of state-specific edges when supported by the data \cite{WangSegarraUhler2020}. Shared edges can be interpreted as conserved circuitry, while state-specific edges provide candidate rewiring hypotheses for downstream validation.
\end{enumerate}

Users may select the causal discovery method that best matches domain assumptions, data modality, and expected mechanism changes. As a guideline, COAST recommends using a \emph{shared-backbone} regime when interactions are expected to be stable and differences arise primarily from mechanism/parameter shifts, and using a \emph{state-specific} regime when genuine wiring changes are expected between $\mathcal{S}$ and $\mathcal{T}$ \cite{WangSegarraUhler2020}. Regardless of the chosen regime, background knowledge (required/forbidden edges, ordering constraints, restricted candidate parent sets) may be incorporated to improve identifiability and biological interpretability \cite{Tetrad}. Before downstream causal modeling and optimization, COAST applies multiple graph falsification (refutation) tests to assess whether the learned structure(s) are statistically compatible with the observed data, using the DoWhy causal modeling framework \cite{dowhy}.

\subsection{Causal Modeling}\label{Causal Modeling}

After structure learning, COAST fits state-specific structural causal models (SCMs) for $\mathcal{S}$ and $\mathcal{T}$ using the learned parent sets. For each state $c\in\{\mathcal{S},\mathcal{T}\}$, let $\mathcal{G}^c$ denote the learned graph (with $\mathcal{G}^{\mathcal{S}}=\mathcal{G}^{\mathcal{T}}$ under a shared backbone -- see Section \ref{Causal Discovery}), and let $\operatorname{pa}_c(i)$ denote the parents of node $i$ in $\mathcal{G}^c$. The induced state-specific joint distribution then factorizes as
\begin{equation}
\mathcal{P}^c(x^c)=\prod_{i=1}^{\mathrm{q}} P^c\!\left(x_i^c \mid x_{\operatorname{pa}_c(i)}^c\right),
\end{equation}
where $x_i^c$ denotes the realization of the same underlying variable $x_i$ in state $c\in\{\mathcal{S},\mathcal{T}\}$, and each conditional distribution $P^{c}(x_i^{c} \mid x_{\operatorname{pa}_{c}(i)}^{c})$ corresponds to a state‑specific causal mechanism. Structural causal models characterize how each variable is generated from its direct causes and an exogenous disturbance term \cite{Pearl2009}. 

Given learned graph structure(s) $\mathcal{G}^{\mathcal{S}}$ and $\mathcal{G}^{\mathcal{T}}$, we assume that each variable $x_i^{c}$, for $i\in\{1,\dots,\mathrm{q}\}$, is generated according to
\begin{equation}
x_i^{c}\;=\; \sum_{j \in \operatorname{pa}_{c}(i)} f_{ij}^{c}\!\left(x_j^{c};\,\zeta_{ij}^{c}\right) \;+\; U_i^{c},
\label{scm-functional}
\end{equation}
where $\operatorname{pa}_{c}(i)$ denotes the parent set of node $i$ in $\mathcal{G}^{c}$, $f_{ij}^{c}(\cdot;\zeta_{ij}^{c})$ is a (potentially nonlinear) function encoding the causal influence of parent variable $x_j^{c}$ on $x_i^{c}$ in state $c$, and $\zeta_{ij}^{c}$ are the corresponding state-specific parameters. The term $U_i^{c}$ represents an exogenous disturbance, which may follow a Gaussian or non-Gaussian distribution.

An SCM associated with $\mathcal{G}^{c}$ encodes both functional and probabilistic assumptions about the underlying data-generating process in state $c$. Under these assumptions, interventional quantities such as $P(y^{c}\mid \mathrm{do}(x^{c}))$ are identifiable via Pearl's do-calculus and related graphical criteria whenever the effect is identifiable in $\mathcal{G}^{c}$ \cite{Pearl2009}. More broadly, SCMs enable principled interventional simulation, and mechanism-level attribution grounded in the learned structural equations and noise terms.

In COAST, the causal modeling module takes as input the inferred graph structure(s) $\mathcal{G}^{\mathcal{S}}$ and $\mathcal{G}^{\mathcal{T}}$ together with the corresponding datasets and fits state-specific mechanisms $f_{ij}^{c}$ for $c\in\{\mathcal{S},\mathcal{T}\}$. The resulting causal models $\mathcal{M}^{c}$, together with the source and target datasets ($D^{\mathcal{S}}$ and $D^{\mathcal{T}}$), are subsequently passed to the \emph{Root Cause Analysis} module.

\subsection{Root Cause Analysis}\label{RCA}

This module identifies and ranks the mechanisms that explain why the system’s distribution differs between the source and target states. We adopt the mechanism-change attribution framework of Budhathoki \emph{et al.}~\cite{Budhathoki2021}, which explains distribution shifts by attributing them to changes in node-wise conditional distributions (``causal mechanisms'') in a probabilistic causal model consistent with the learned DAG. 

\subsubsection{Attributing changes in distributions}
The central idea of this attribution method is to explain the observed distributional shift between the source and target states in terms of changes in individual causal mechanisms. Given state-specific causal models $\mathcal{M}^{\mathcal{S}}$ and $\mathcal{M}^{\mathcal{T}}$ learned from $D^{\mathcal{S}}$ and $D^{\mathcal{T}}$, respectively, the method progressively replaces mechanisms from the source model with their counterparts inferred in the target model.

Concretely, let $\Gamma \subseteq \{1,\ldots,\mathrm{q}\}$ denote a set of nodes whose mechanisms have been replaced. For a given node $i$, define the hybrid interventional distribution obtained by using target-state mechanisms for nodes in $\Gamma$ and source-state mechanisms for all other nodes:
\[
P_\Gamma(x_i)
\;=\;
\sum_{x_{-i}}
\prod_{j \in \Gamma}
P^{\mathcal{T}}\!\left(x_j^{\mathcal{T}} \mid x^{\mathcal{T}}_{\operatorname{pa}_{\mathcal{T}}(j)}\right)
\prod_{j \notin \Gamma}
P^{\mathcal{S}}\!\left(x_j^{\mathcal{S}} \mid x^{\mathcal{S}}_{\operatorname{pa}_{\mathcal{S}}(j)}\right),
\]
where $x_{-i}$ denotes all variables except $x_i$, and the distributions on the right-hand side are induced by the corresponding structural equations in $\mathcal{M}^{\mathcal{S}}$ and $\mathcal{M}^{\mathcal{T}}$.
Replacing mechanisms that differ across states alters downstream marginal distributions, whereas replacing unchanged mechanisms leaves them invariant.

To fairly attribute the overall distributional shift to individual nodes, the method employs Shapley symmetrization \cite{Shapley1953,Budhathoki2021}, which averages marginal contributions over all possible orders in which mechanisms are replaced. For each node $i$, its attribution score is computed by comparing the marginal distribution of $x_i$ with and without replacing node $j$'s mechanism, averaged across all subsets $\Gamma \subseteq \{1,\ldots,\mathrm{q}\}\setminus\{j\}$.

The discrepancy between marginals can be quantified using any suitable distance measure. While prior work commonly employs information-theoretic divergences (e.g., KL divergence), in COAST we quantify distributional change using differences in node-wise means. This choice aligns naturally with our transition percentage metric (Section \ref{trans_pct}) and yields a stable, interpretable measure of state transition in the settings considered. Importantly, the attribution framework is agnostic to the specific choice of discrepancy measure, which only specifies how distributional change is quantified.

Finally, we emphasize that a change in a causal mechanism is interpreted in the standard SCM sense: a change in the conditional distribution governing a variable given its parents. Such changes may arise from differences in the functional form $f_{ij}^c$, from parameter shifts $\zeta_{ij}^c$, or from changes in the associated disturbance term $U_i^c$. From a causal attribution perspective, these effects are equivalent, as all correspond to state-specific differences in the mechanism generating $x_i^c$ and thus contribute to the observed source-to-target distributional shift.

\subsubsection{Calculating cumulative attribution scores}
To quantify the contribution of each node in the causal graph to the overall distributional shift between the source and target states, we compute accumulated attribution scores by aggregating node-level Shapley-based attributions across variables exhibiting statistically significant marginal shifts between the two conditions. For each such variable, we apply the distribution change attribution method described above to obtain node‑specific attributions. To derive a global measure of each node’s influence across the full set of responsive variables, we aggregate these contributions by summing the absolute values of its Shapley‑based attribution scores. Formally, let $\psi_{ij}$ denote the attribution score of node $j$ with respect to variable $i$. The accumulated attribution score for node $j$ is defined as

\begin{equation}
\label{eq:attr_score}
\mathrm{u}_j = \sum_{i \in [\omega]} |\psi_{ij}|,
\end{equation}
where $[\omega]$ denotes the set of variables exhibiting statistically significant marginal shifts between the source and target states. Taking absolute values prevents cancellations between contributions of opposite sign and captures the total magnitude of each node’s influence on the observed distributional changes. The resulting accumulated scores are used to rank and prioritize candidate intervention targets.

\subsubsection{Determining the set of candidate intervention targets}
To determine the number of candidate intervention targets from the ranked list of accumulated attribution scores, we employed an adaptive selection procedure based on a diminishing-returns criterion. The accumulated attribution scores were first normalized so that they sum to 100, representing each node's percentage contribution to the total attribution. Starting from the highest-scoring node, candidates were added sequentially in descending order of their normalized scores. At each step, the normalized score of the next candidate was compared to the cumulative sum of the scores of all previously selected candidates. The selection process terminated when the ratio of the next candidate's score to the cumulative score fell below a predefined threshold $\delta$ (e.g., $\delta = 0.01$), indicating that the marginal contribution of additional candidates had become negligible relative to the already-selected set. This adaptive stopping rule avoids the need for a fixed, arbitrary cutoff on the number of intervention targets, and instead selects a parsimonious set of nodes that collectively account for the dominant share of the total attribution.

\subsection{Optimal Intervention Identifier}\label{OpIntIdent}

The Optimal Intervention Identifier converts causal insights into actionable intervention strategies that drive the system from a source to a target state. Given influential drivers identified by root cause analysis, this module determines both the identities and magnitudes of interventions while enforcing feasibility, stability, and sparsity constraints. COAST supports two operating modes: a \emph{hypothesis‑agnostic} mode, in which intervention targets and their cardinality are inferred directly from data, and a \emph{hypothesis‑guided} mode, in which prior knowledge constrains the intervention size and/or admissible targets.
\subsubsection{Hypothesis‑Agnostic Sparse Optimal Intervention Design}\label{hypAgn}

The Optimal Intervention Identifier module takes as input a set of candidate intervention targets, such as the $\mathrm{k}$ most influential drivers of change identified by the root cause analysis (Section~\ref{RCA}). Its goal is to select \emph{parsimonious} subsets of these drivers and determine corresponding intervention values that induce an effective transition from the source state to the target state.

An intervention is represented by a vector $\alpha=(\alpha_1,\ldots,\alpha_\mathrm{k})\in\mathbb{R}^{\mathrm{k}}$ and is defined as a modification of the causal mechanisms generating a subset of variables. Specifically, interventions act by altering the conditional distributions

\begin{equation}
P^{c}\!\left(x_i^{c} \mid x_{\operatorname{pa}_{c}(i)}^{c}\right)
\;\longrightarrow\;
\dot{P}^{c}\!\left(x_i^{c} \mid x_{\operatorname{pa}_{c}(i)}^{c}\right),
\qquad i=\lbrace1,\ldots,\mathrm{q}\rbrace,\; c\in\{\mathcal{S},\mathcal{T}\},
\end{equation}
where the dot notation indicates quantities evaluated under the post‑intervention regime.

Any variable in index set $[m]\subseteq[k]\subseteq[q]$, where $\alpha_i\neq 0,\ \forall\, i\in[m]$ is referred to as an \emph{interventional target}. Without loss of generality, we focus on \emph{shift interventions}~\cite{sen2017,zhang2021}, a special class of \emph{soft interventions}~\cite{eberhardt2007}. Under a shift intervention, the structural assignment in Eq.~(\ref{scm-functional}) is modified as
\begin{equation}
\label{eq:shift-intervention}
\dot{x}_i^{c}
\;=\;
\sum_{j\in\operatorname{pa}_{c}(i)} f_{ij}^{\,c}\!\left(x_j^{c};\zeta_{ij}^{\,c}\right)
\;+\;
\alpha_i
\;+\;
U_i^{\,c},
\qquad \forall\, i\in[m],\; c\in\{\mathcal{S},\mathcal{T}\}.
\end{equation}

Determining whether an intervention achieves the desired outcome relies on samples drawn from the intervened source regime, denoted $\dot{x}^{\mathcal{S}}\sim \dot{P}^{\mathcal{S}}$. Since $\dot{x}^{\mathcal{S}}$ is a random vector, we estimate relevant expectations (e.g., distributional means or other summary statistics) using empirical averages over Monte-Carlo samples~\cite{sen2017,zhang2021}.

\paragraph{Constrained multi-objective optimization design}

We introduce a novel unified constrained multi‑objective optimization formulation for causal intervention design under distribution shift, which jointly accounts for source‑to‑target alignment, target‑state stability, and sparsity of interventions.

COAST formulates the search for optimal interventions as a constrained multi-objective optimization that handles conflicting goals and incorporates fixed as well as custom constraints.

\begin{equation}
\label{main-objective}
\arg\min_{\alpha \in \mathbb{R}^{\mathrm{k}}}
\;
\underbrace{\left\|\,\dot{\bar{x}}^{\mathcal{S}} - \bar{x}^{\mathcal{T}}\,\right\|_{2}^{2}}_{\text{source}\rightarrow\text{target alignment}}
\;+\;
\gamma\,\underbrace{\left\|\,\dot{\bar{x}}^{\mathcal{T}} - \bar{x}^{\mathcal{T}}\,\right\|_{2}^{2}}_{\text{target-state stability}}
\;+\;
\lambda\,\underbrace{\sum_{i=1}^{\mathrm{k}} \mathrm{w}_i \lvert \alpha_i\rvert}_{\text{weighted }\ell_1\text{ sparsity}}.
\end{equation}

subject to a set of user-defined, feasibility and stability constraints:
\begin{equation*}
\begin{aligned}
&\textbf{(C1) Range constraints:} \qquad
\dot{\bar{x}}_i^{\mathcal{S}},\, \dot{\bar{x}}_i^{\mathcal{T}} \in [\mathrm{min},\,\mathrm{max}],
\quad \forall i \in \{1,\dots,\mathrm{q}\}, \\[0.6em]
&\textbf{(C2) Actionable variables:} \qquad
\alpha_i = 0, \quad \forall\, i \notin [m] \subseteq [k] \subseteq [q] \\[0.6em]
&\textbf{(C3) Relative-change preservation:} \qquad
\left|\frac{\dot{\bar{x}}_i^{\mathcal{S}}}{\bar{x}_i^{\mathcal{S}}}\right| \le \mathrm{RC}, \qquad
\left|\frac{\dot{\bar{x}}_i^{\mathcal{T}}}{\bar{x}_i^{\mathcal{T}}}\right| \le \mathrm{RC},
\quad \forall i \in \{1,\dots,\mathrm{q}\}.
\end{aligned}
\end{equation*}

Here, $\bar{x}^{\mathcal{S}},\bar{x}^{\mathcal{T}}\in\mathbb{R}^{\mathrm{q}}$ denote the empirical means under the unperturbed source and target environments, respectively, and $\dot{\bar{x}}^{\mathcal{S}},\dot{\bar{x}}^{\mathcal{T}}\in\mathbb{R}^{\mathrm{q}}$ the corresponding means under intervention. The discrepancy between $\dot{\bar{x}}^{\mathcal{S}}$ and $\bar{x}^{\mathcal{T}}$ defines the primary objective to be minimized.\\

Parameters are defined as follows:
\begin{itemize}
    \item $\mathrm{w}_i \in [0,1]$ denote normalized feature weights derived from the reciprocals of the accumulated attribution scores $\mathrm{u}_i$ (see Eq.~(\ref{eq:attr_score})).
    \item $\gamma \in [0,1]$ is user‑defined and application‑dependent parameter, reflecting the relative importance assigned to target‑state stability versus source‑to‑target alignment.
    \item $\lambda \in \mathbb{R}$ regulates sparsity of the intervention vector $\alpha$.
    \item $\mathrm{RC} \in [0,\mathrm{max}]$ bounds allowable relative changes.
\end{itemize}

For interpretation of constraints:

\noindent \textbf{(C1)} enforces physically valid, operable, or regulation‑compliant ranges post‑intervention. This prevents non-physiological or model-invalid values e.g. negative gene expression levels, extreme activity outside assay-defined ranges etc.

\noindent \textbf{(C2)} limits interventions to a (controllable/permitted) subset $[m]$ of the $\mathrm{k}$ most influential drivers of change identified by the root cause analysis.


\noindent \textbf{(C3)} caps relative deviations to preserve critical behaviors, such as safety margins and application‑specific constraints, while still enabling progress toward the target state.
\\

The proposed formulation (see eq.(\ref{main-objective})) integrates three complementary terms that jointly balance \emph{alignment}, \emph{stability}, and \emph{sparsity}:

\begin{itemize}
    \item \emph{Alignment term} $\left\|\,\dot{\bar{x}}^{\mathcal{S}} - \bar{x}^{\mathcal{T}}\,\right\|_{2}^{2}$ encourages the shifted source mean to approach the target mean, reducing distributional divergence between intervened source and target states.
    \item \emph{Target-state stability term} 
$\gamma \left\|\, \dot{\bar{x}}^{\mathcal{T}} - \bar{x}^{\mathcal{T}} \,\right\|_{2}^{2}$ penalizes deviations from the original target-state, thereby encouraging post-intervention stability in the target environment. 
When $\gamma = 0$, the optimization ignores target-state homeostasis entirely. As $\gamma$ increases, deviations in the target environment become more heavily penalized. This term is particularly important in reversibility settings (e.g., steering diseased cells toward healthy phenotypes while ensuring that healthy cells remain minimally perturbed).
    \item \emph{Weighted sparsity term} $\lambda\sum_{i=1}^{k} \mathrm{w}_i |\alpha_i|$ promotes parsimonious interventions, focusing changes on the most influential variables (via $\mathrm{w}_i$) to reduce complexity and enhance interpretability.
\end{itemize}

The optimization in eq.(\ref{main-objective}) is typically solved along a grid of $\lambda\in\mathbb{R}$ to regulate sparsity (and hence the number of interventional targets). Next we present the method used to determine the range of $\lambda$ values.

\paragraph{Determining the Range of $\boldsymbol{\lambda}$ for Sparse-to-Dense Solutions}

Given the objective in eq.(\ref{main-objective}), we determine a principled range for the regularization parameter $\lambda$ using the Karush--Kuhn--Tucker (KKT) conditions for weighted $\ell_1$-penalized optimization \cite{boyd2004convex,tibshirani1996lasso}. Let $L(\alpha)$ denote the smooth part of the objective, comprising the source-to-target alignment and target stability terms. We consider the fully sparse candidate solution $\alpha=\mathbf{0}$ and evaluate the stationarity condition at this point.

Under the additive intervention model implicit in eq.(\ref{main-objective}), the perturbed source and target centroids satisfy $\dot{\bar{x}}^{\mathcal{S}} = \bar{x}^{\mathcal{S}} + \alpha$ and $\dot{\bar{x}}^{\mathcal{T}} = \bar{x}^{\mathcal{T}} + \alpha$, respectively, so that the corresponding Jacobians with respect to $\alpha$ are identity matrices. The gradient of the smooth loss evaluated at $\alpha=\mathbf{0}$ therefore reduces to
\[
g \;=\; \nabla_{\alpha} L(\alpha)\big|_{\alpha=\mathbf{0}}
\;=\;
2\big(\bar{x}^{\mathcal{S}} - \bar{x}^{\mathcal{T}}\big).
\]

For each coordinate $i$, the subgradient of $|\alpha_i|$ is \cite{tibshirani1996lasso,boyd2004convex}:
\[
\mathrm{u}_i =
\begin{cases}
\operatorname{sign}(\alpha_i) \cdot \mathrm{w}_i, & \alpha_i \neq 0, \\
\in [-\mathrm{w}_i, \mathrm{w}_i], & \alpha_i = 0.
\end{cases}
\]

This means:
\begin{itemize}
    \item If $\alpha_i \neq 0$, the subgradient is fixed at $\pm \mathrm{w}_i$.
    \item If $\alpha_i = 0$, the subgradient can take any value in $\left[-\mathrm{w}_i, \mathrm{w}_i\right]$.
\end{itemize}

At the fully sparse solution ($\alpha_i = 0$ for all $i$):
\[
\mathrm{g}_i = \left.\frac{\partial L(\alpha)}{\partial \alpha_i}\right|_{\alpha=\mathbf{0}}.
\]
The KKT stationarity condition requires:
\[
\mathrm{g}_i + \lambda \mathrm{u}_i= 0, \quad \mathrm{u}_i \in [-\mathrm{w}_i, \mathrm{w}_i].
\]

For $\mathrm{u}_i$ to be in $\left[-\mathrm{w}_i, \mathrm{w}_i\right]$, we need:
\[
-\frac{\mathrm{g}_i}{\lambda} \in [-\mathrm{w}_i, \mathrm{w}_i] \;\Longrightarrow\; |\mathrm{g}_i| \le \lambda \mathrm{w}_i.
\]
If $|\mathrm{g}_i| > \lambda \mathrm{w}_i$, then even the largest possible subgradient ($\pm \mathrm{w}_i$) scaled by $\lambda$ cannot cancel $\mathrm{g}_i$, so $\alpha_i = 0$ cannot be optimal.\\

To guarantee all coefficients remain zero \cite{tibshirani1996lasso,friedman2010glmnet}:
\[
\lambda_{\max} = \max_i \frac{|\mathrm{g}_i|}{\mathrm{w}_i}.
\]
For any $\lambda \ge \lambda_{\max}$, the zero solution satisfies KKT and is optimal. For smaller $\lambda$, coefficients begin to enter the model, creating a path from sparse to dense solutions \cite{friedman2010glmnet}.\\

Then, we define:
\[
\lambda_{\min} = \epsilon \cdot \lambda_{\max}, \quad \epsilon \in [10^{-3}, 10^{-4}],
\]
and generate a logarithmic sequence of $\Theta$ values between $\lambda_{\max}$ and $\lambda_{\min}$:
\[
\lambda_\theta = \lambda_{\max} \left( \frac{\lambda_{\min}}{\lambda_{\max}} \right)^{\frac{\theta-1}{\Theta-1}}, \quad \theta= \lbrace 1,\ldots,\Theta \rbrace.
\]

This approach avoids heuristic choices and guarantees coverage from fully sparse to nearly dense solutions. When weights $\mathrm{w}_i$ are derived from attribution scores, the penalty becomes context-aware, prioritizing interventions on less influential variables and delaying adjustments to highly influential ones unless $\lambda$ is sufficiently small. This causally informed regularization path improves interpretability and aligns intervention sparsity with domain knowledge \cite{tibshirani1996lasso,zou2006adaptive}. Collectively, this construction yields a controlled regularization path $\{\alpha(\lambda)\}$ that enables systematic exploration of parsimonious versus more complex intervention sets.

\paragraph{Persistence along the regularization path}

Let $\ell=\{\lambda_1,\dots,\lambda_\mathrm{L}\}$ denote the set of sparsity (regularization) parameters evaluated. For each $\lambda\in\ell$, solving the intervention optimization problem yields a candidate solution in the form of an intervention set $\mathcal{I}_\lambda$ (that is, the subset of variables selected for intervention). Depending on $\lambda$, $\mathcal{I}_\lambda$ may contain a single target or multiple targets, reflecting different sparsity-effectiveness trade‑offs along the regularization path.

We define the persistence of a candidate intervention set $\mathcal{I}$ as
\begin{equation}
\label{eq:stability-score}
\mathrm{Pers}(\mathcal{I})
\;=\;
\frac{1}{|\ell|}
\sum_{\lambda\in\ell}
\mathbf{1}\!\left\{\mathcal{I}_\lambda=\mathcal{I}\right\},
\end{equation}
where $\mathbf{1}\{\cdot\}$ is the indicator function.

$\mathrm{Pers}(\mathcal{I})$ quantifies robustness to the choice of sparsity parameter by measuring how frequently the \emph{same single- or multi-target intervention set} reappears across the solutions $\{\mathcal{I}_\lambda\}_{\lambda\in\ell}$.

Analogously, we define the persistence for a single variable $x_i$ as
\begin{equation}
\label{eq:target-stability}
\mathrm{Pers}(x_i)
\;=\;
\frac{1}{|\ell|}
\sum_{\lambda\in\ell}
\mathbf{1}\!\left\{x_i\in\mathcal{I}_\lambda\right\},
\end{equation}
which captures how consistently $x_i$ is selected \emph{as a member of the candidate intervention sets} across sparsity levels. In contrast to $\mathrm{Pers}(\mathcal{I})$, which assesses persistence of an entire combination, $\mathrm{Pers}(x_i)$ measures the marginal consistency of individual targets aggregated over all candidate solutions.

Intervention sets and variables that persist over a broad range of $\lambda$ values are less sensitive to regularization tuning and thus reflect robust trade-offs between transition effectiveness and sparsity. While persistence does not imply optimality, it provides a principled robustness criterion that complements objective value and supports prioritization of intervention strategies for downstream evaluation and experimental validation.

\paragraph{Transition percentage}\label{trans_pct}
To evaluate the effectiveness of a given intervention in driving the system from a source state toward a target state, we define a normalized metric termed the \emph{transition percentage}. Let $\bar{x}^{\mathcal{S}} = (\bar{x}^{\mathcal{S}}_1,\ldots,\bar{x}^{\mathcal{S}}_\mathrm{q})$ and $\bar{x}^{\mathcal{T}} = (\bar{x}^{\mathcal{T}}_1,\ldots,\bar{x}^{\mathcal{T}}_\mathrm{q})$ denote the vectors of node-wise sample means under the source and target states, respectively, where $\mathrm{q}$ is the number of variables (nodes) in the causal graph. The total baseline distance between the two states is defined as
\[
\mathrm{d}_{\mathrm{total}} = \sum_{i=1}^{\mathrm{q}} \left(\bar{x}^{\mathcal{S}}_i - \bar{x}^{\mathcal{T}}_i\right)^2.
\]

After applying an intervention to a selected subset of nodes, samples are generated from the intervened causal model, yielding node-wise means $\dot{\bar{x}} = (\dot{\bar{x}}_1,\ldots,\dot{\bar{x}}_\mathrm{q})$. The residual distance from the target state is then
\[
\mathrm{d}_{\mathrm{res}} = \sum_{i=1}^{\mathrm{q}} \left(\dot{\bar{x}}^{\mathcal{S}}_i - \bar{x}^{\mathcal{T}}_i\right)^2.
\]

The transition percentage is defined as
\begin{equation}
\label{TP}
s = \left(1 - \frac{\mathrm{d}_{\mathrm{res}}}{\mathrm{d}_{\mathrm{total}}}\right)\times 100\%.
\end{equation}

This metric quantifies the fraction of the total source-to-target shift that is explained by the intervention. A value of $100\%$ indicates that the intervention fully recapitulates the target state at the level of node-wise means, whereas a value of $0\%$ indicates no progress relative to the source state. Intermediate values reflect partial transitions toward the target state. By normalizing against the baseline source--target distance, the transition percentage enables consistent comparison of intervention effectiveness across different experimental settings and scales.

\subsubsection{Hypothesis‑Guided Fixed‑Cardinality Intervention Design}
The regularized formulation described above (Section~\ref{hypAgn}) supports \emph{hypothesis‑agnostic} discovery of sparse intervention sets when the appropriate level of intervention sparsity is not specified \emph{a priori}. In many practical settings, however, domain knowledge or operational constraints restrict the intervention space for example, when only a curated set of targets is considered actionable, or when no more than a fixed number of targets can be perturbed simultaneously. COAST accommodates such \emph{hypothesis‑guided} scenarios by constraining the search to a user‑specified candidate set and enforcing a fixed intervention cardinality.

Let $[z]\subseteq[q]$ denote a set of candidate intervention targets derived from prior knowledge, and let $\mathrm{k}$ be the prescribed intervention cardinality. The objective is to identify a subset $[\epsilon]\subseteq[z]$ with $|[\epsilon]|=\mathrm{k}$ that maximizes the source‑to‑target transition objective. Exhaustive evaluation of all $\binom{|[z]|}{\mathrm{k}}$ candidate combinations is typically computationally infeasible; COAST therefore employs a two‑stage prioritization strategy to focus computation on the most promising subsets.

\textbf{Stage 1 (single‑target screening):} For each candidate target node $i \in[z]$, we solve the transition‑optimization problem in Eq.~(\ref{main-objective}) with the sparsity regularizer removed and the intervention restricted to target node $i$ only. This yields an optimal single‑target intervention magnitude $\alpha_i$ together with an associated transition percentage $s_i$ (see Eq.~\ref{TP}), which quantifies the impact of intervening on target node $i$.

\textbf{Stage 2 (combination prioritization):} Candidate $\mathrm{k}$‑target subsets are ranked using an aggregate heuristic based on their constituent single‑target scores. In this work, we use the mean score
\begin{equation}
\bar{s}([\epsilon])=\frac{1}{|[\epsilon]|}\sum_{i\in[\epsilon]} s_i,
\end{equation}
though alternative aggregation schemes can be employed. This screening strategy is motivated by the observation that the transition objective typically exhibits strong concentration around high‑impact targets: single‑target screening provides a low‑cost surrogate for each variable’s marginal causal contribution, enabling efficient elimination of ``weak'' combinations. Under mild assumptions of approximate additivity or diminishing returns, targets with strong individual effects are likely to appear in near‑optimal multi‑target solutions. Consequently, ranking combinations by aggregated single‑target scores concentrates evaluation on a small region of the combinatorial search space that, with high probability, contains the most effective state‑transition solutions.

COAST subsequently evaluates top ranked combinations using the unregularized transition objective (without the sparsity term) and returns the optimal intervention values, subject to the constraint that interventions are permitted only on the variables in $[\epsilon]$. This two‑stage strategy substantially reduces the number of expensive objective evaluations while preserving a direct and principled connection to the underlying causal optimization problem.

\subsection{Evaluation of Interventions}\label{DomainEval}

This module evaluates the solutions proposed by the Optimal Intervention Identifier and provides multiple layers of interpretability. First, it summarizes solution characteristics through visual analytics. A persistence diagram quantifies the persistence of each candidate intervention by reporting how consistently specific intervention sets or individual variables reoccur across the regularization path. Additional plots illustrate how the number of intervention targets varies as the regularization parameter $\lambda$ changes, together with the corresponding transition percentage achieved by each solution. These visualizations facilitate a clear assessment of the trade‑offs between intervention sparsity, transition performance, and robustness with regard to regularization tuning. Finally, a heatmap depicts the intervention intensity associated with each influential variable across all candidate solutions, enabling fine‑grained comparison of intervention magnitudes.

\emph{Biomedical instantiation:} To deepen biological insights, COAST enable us to perform in-silico simulations of the proposed interventions using the learned causal model. Post-intervention, we identify genes with significantly altered expression and conduct pathway enrichment analysis leveraging resources such as Gene Ontology (GO), Kyoto Encyclopedia Genes Genomes (KEGG), and Gene Set Enrichment Analysis (GSEA) \cite{go,kegg,gsea}. This analysis highlights biological pathways modulated by the interventions. Finally, the pipeline compiles pathway analysis results, including visualizations and statistical metrics, to facilitate interpretation and evaluation of the biological impact of each intervention strategy.

For the disease reversibility application, our pipeline further integrates with the Open Targets platform via its API \cite{opentargets}. For each identified target, we retrieve disease-association scores, which offer additional context to support interpretation and prioritization of subsequent experimental validation.

\backmatter

\paragraph{Acknowledgements}
We thank Dr. Michael Kavana for reviewing the manuscript, providing valuable feedback, and for his continued support and encouragement in pursuing and implementing this work.

\section*{Declarations}

 \paragraph{Conflict of interest/Competing interests.} 
 Z.S., U.M., and D.V.M. are employees of Merck Sharp \& Dohme LLC, a subsidiary of Merck \& Co., Inc., Rahway, NJ, USA, and may hold Merck \& Co., Inc., Rahway, NJ, USA stock.

\paragraph{Ethics approval and consent to participate}
Not applicable.

\paragraph{Consent for publication}
Not applicable

\paragraph{Data availability} 
The data supporting the findings of this study will be made publicly available upon publication of the manuscript in a peer‑reviewed journal.

\paragraph{Materials availability}
All materials associated with this study will be made available upon publication of the manuscript in a peer‑reviewed journal.

\paragraph{Code availability}
The code used to implement the COAST pipeline will be released upon publication of the manuscript in a peer‑reviewed journal.

\paragraph{Author contribution} 
Z.S. and D.V.M. developed and implemented the COAST framework. Z.S. and D.V.M. designed the experiments. Z.S. performed the experiments, analyzed the data, and integrated the resulting analyses into the manuscript. Z.S. and D.V.M. interpreted the results and wrote the manuscript. U.M. contributed to the experimental design, reviewed the manuscript, and provided critical feedback and comments. D.V.M. conceptualized, initiated, and supervised the study.


\newpage
\begin{appendices}

\section{Average transition percentage}\label{secA1}

As COAST identifies a series of solutions with different numbers of intervention targets using different values of the regularization strength $\lambda$, average transition percentage is defined as the mean of transition percentages over all non-zero solutions along the regularization path. It can represent the overall performance of COAST across a range of intervention solutions.

\begin{figure}[H]
  \centering
  \includegraphics[width=\linewidth]{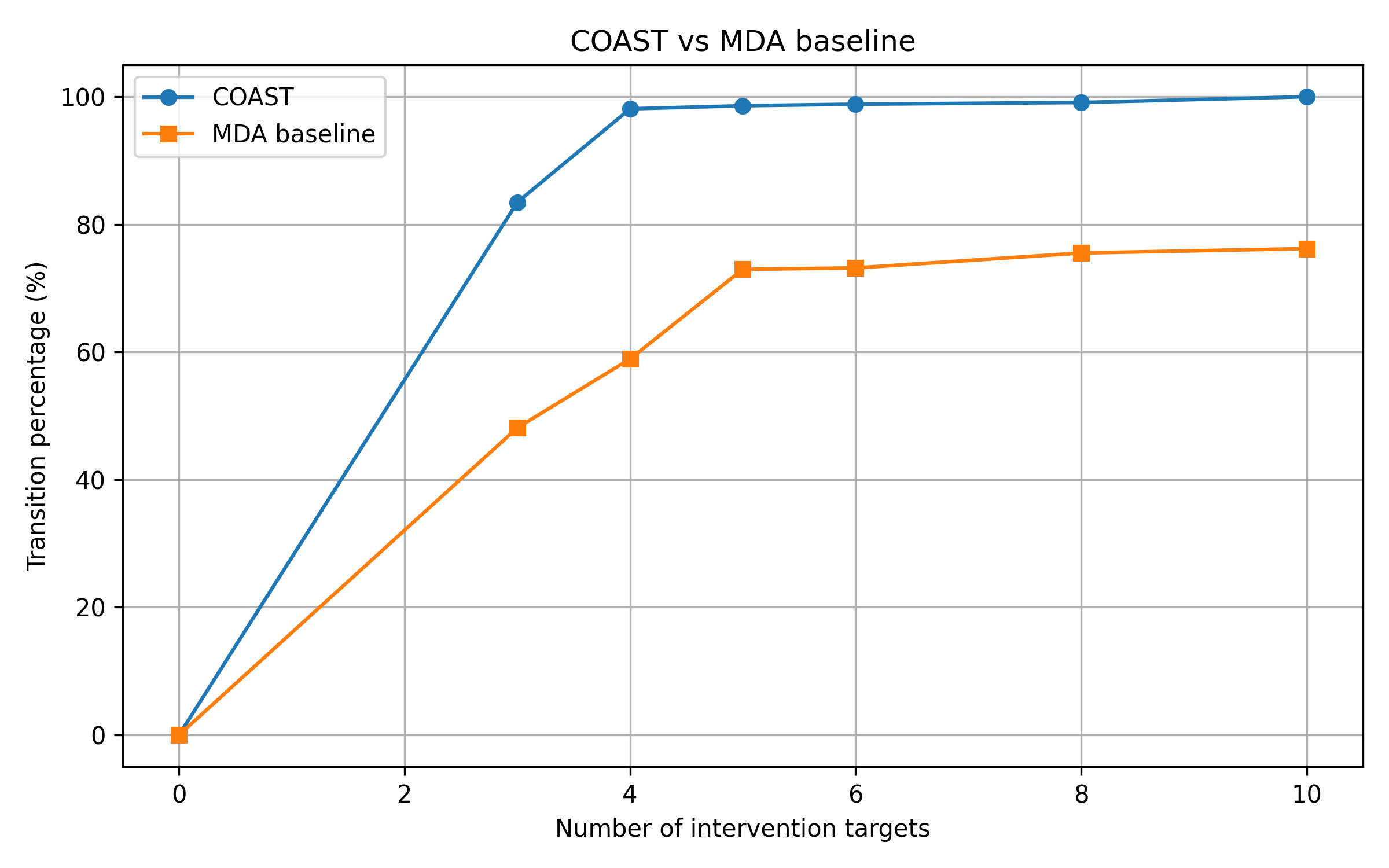}
  \caption{Comparison of COAST and the MDA baseline on the transition percentages along the regularization path. Average transition percentage is calculated as the mean of the transition percentages at different number of intervention targets except 0.}
  \label{fig:avg_trans_pct}
\end{figure}

To compare COAST with the MDA baseline, we generated transition percentages of MDA using the same sizes of interventions, i.e., the top $k$ variables with the largest MDA scores, with $k$ matched to the numbers of intervention targets identified by COAST on its regularization path (Fig. \ref{fig:avg_trans_pct}).

\end{appendices}

\clearpage
\section*{Supplementary Information}

\subsection*{Supplementary results on synthetic experiments}\label{secA2}

We present additional synthetic experiment results for 10-node and 500-node graphs, complementing the 100-node results reported in the main text. The same experimental protocol, evaluation metrics, and baseline comparison are used throughout: five random seeds per configuration, noise levels $\sigma \in \{1, 3, 5\}$, and numbers of intervention targets $k \in \{1, 5, 10\}$.

Figure~\ref{fig:synthetic_results_10} summarizes the comparison between COAST and the MDA baseline on 10-node graphs. In this small-graph regime, both methods achieve strong performance, reflecting the reduced combinatorial complexity of the intervention identification problem. For recall@$k$, COAST achieves perfect recall across all single-target and 10-target settings, with a slight reduction to 0.96 at $k=5$ under medium and high noise. The MDA baseline also performs well, reaching 1.00 in all single-target and 10-target configurations but dropping to 0.92 at $k=5$, yielding a modest advantage of 4--8\% for COAST. Transition percentage@$k$ values are similarly close: COAST ranges from 95\% to 100\% and the MDA baseline from 91\% to 100\%, with both methods achieving near-perfect scores in the 10-target setting where the intervention set spans the full variable space. The average transition percentage follows a comparable pattern, with COAST ranging from 73\% to 100\% and MDA from 70\% to 100\%; here the gap is most apparent in the multi-target settings ($k \in \{5, 10\}$), where COAST leads by 2--5\%. Overall, the small graph size limits the room for differentiation, as both methods can effectively leverage the limited variable space.

Figure~\ref{fig:synthetic_results_500} presents the corresponding results for 500-node graphs, where the substantially larger search space provides a more challenging test of each method's ability to pinpoint the correct intervention targets. COAST maintains excellent recall@$k$ across all settings, ranging from 0.94 to 1.00, while the MDA baseline deteriorates dramatically: recall drops to 0.40 for $k=1$, falls to 0.00 for $k=5$ (indicating complete failure to identify any true target among the top-ranked variables), and recovers only slightly to 0.08 for $k=10$. COAST outperforms MDA by 60--96\% across all conditions, underscoring the critical advantage of causal-model-guided attribution over marginal distributional comparisons as the variable space grows. For transition percentage@$k$, COAST achieves values between 94\% and 100\%, whereas MDA ranges from only 50\% to 66\%, resulting in an absolute improvement of 34--46\%. The average transition percentage further confirms this trend: COAST maintains consistently high values of 94\%--100\%, while MDA falls to 40\%--57\%, with COAST leading by 40--54\%. Notably, COAST's performance remains remarkably stable across noise levels and numbers of targets even in this high-dimensional setting, whereas MDA shows both lower absolute performance and greater sensitivity to the problem configuration.

\setcounter{figure}{0}
\renewcommand{\thefigure}{S\arabic{figure}}

\begin{figure}[H]
  \centering
  \includegraphics[width=\linewidth]{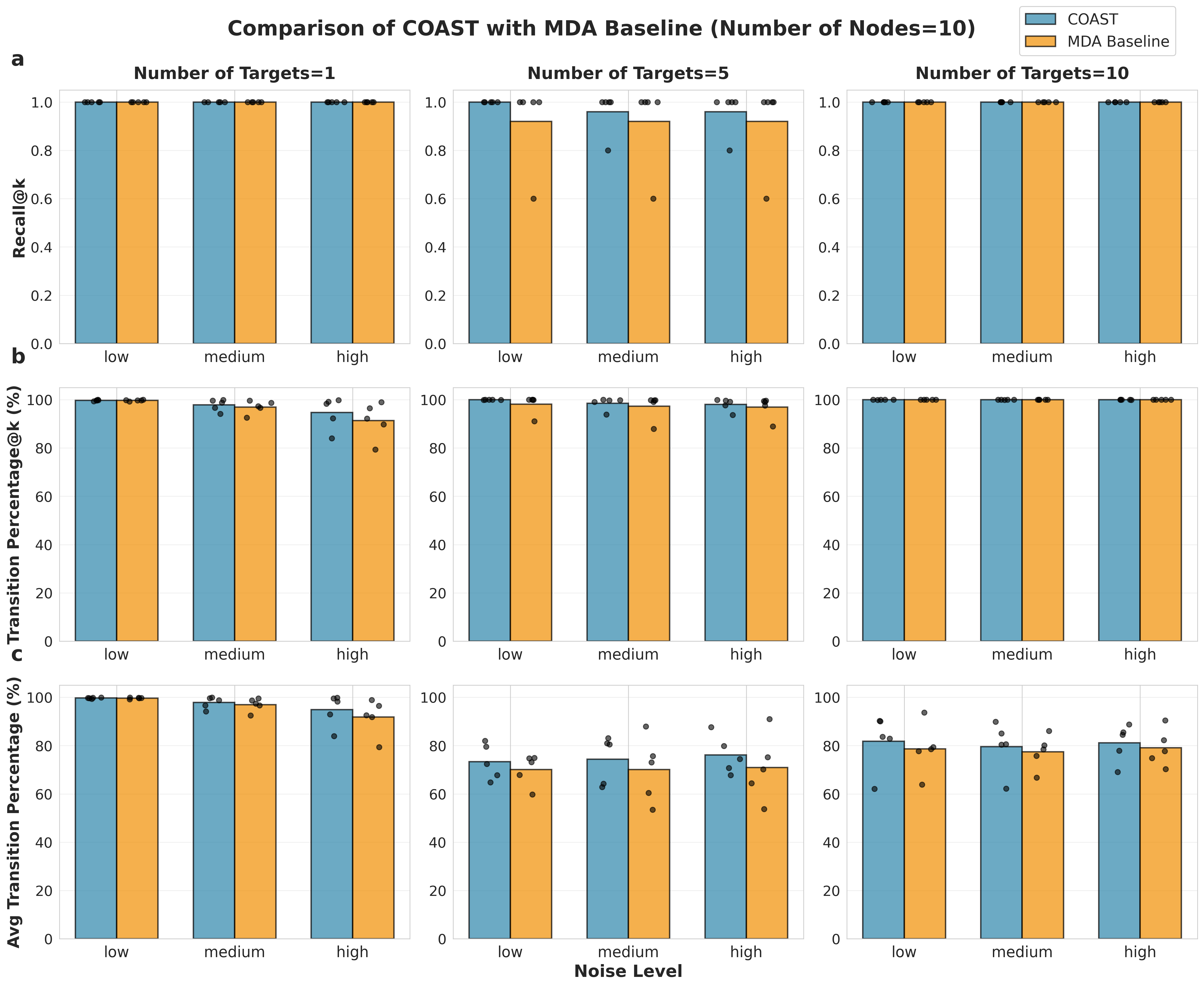}  
  \caption{\textbf{Results of synthetic datasets on 10-node graphs.} Comparison of COAST and the MDA baseline on 10-node graphs across noise levels (low: $\sigma=1$, medium: $\sigma=3$, high: $\sigma=5$). Within each row, the three panels correspond to different numbers of intervention targets ($k \in \{1, 5, 10\}$). Bar heights represent means over five random seeds, and black dots show individual seed values. \textbf{(a)} Recall@$k$. Both methods achieve high recall in this small-graph regime, with COAST maintaining a slight advantage at $k=5$. \textbf{(b)} Transition percentage@$k$. Performance is comparable between methods, with both approaching 100\% in the 10-target setting. \textbf{(c)} Average transition percentage across all non-zero solution sizes along the regularization path. COAST holds a consistent 2--5\% lead in multi-target configurations.}
  \label{fig:synthetic_results_10}
\end{figure}

\begin{figure}[H]
  \centering
  \includegraphics[width=\linewidth]{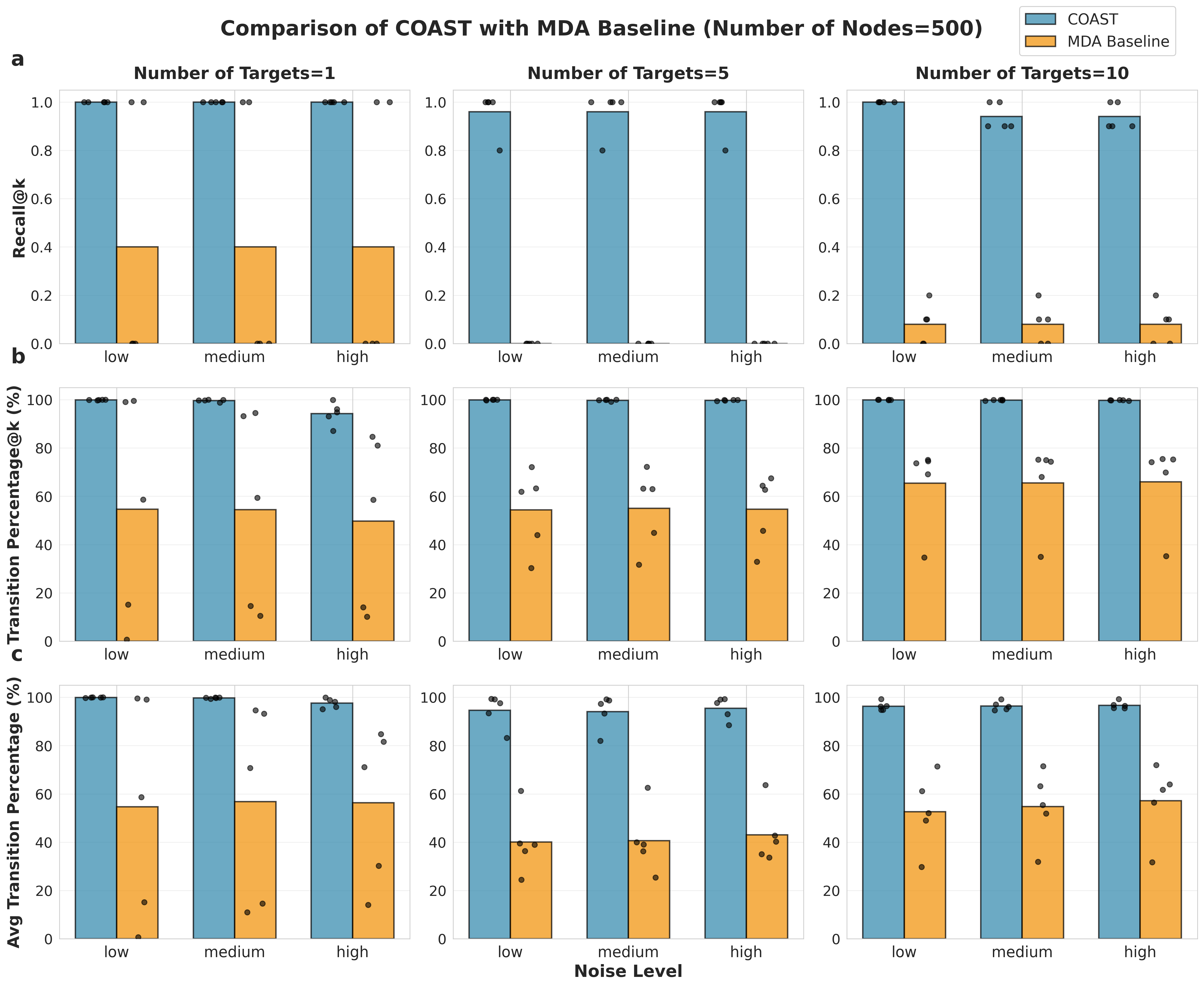}  
  \caption{\textbf{Results of synthetic datasets on 500-node graphs.} Comparison of COAST and the MDA baseline on 500-node graphs across noise levels (low: $\sigma=1$, medium: $\sigma=3$, high: $\sigma=5$). Within each row, the three panels correspond to different numbers of intervention targets ($k \in \{1, 5, 10\}$). Bar heights represent means over five random seeds, and black dots show individual seed values. \textbf{(a)} Recall@$k$. COAST achieves near-perfect recall (0.94--1.00) across all settings, while the MDA baseline drops to 0.00--0.40, demonstrating COAST's scalability advantage. \textbf{(b)} Transition percentage@$k$. COAST outperforms MDA by 34--46\%, with MDA unable to exceed 66\% in any configuration. \textbf{(c)} Average transition percentage across the regularization path. COAST maintains values above 94\% regardless of noise or number of targets, whereas MDA ranges from 40\% to 57\%, confirming the robustness of COAST's causal-model-guided optimization in high-dimensional settings.}
  \label{fig:synthetic_results_500}
\end{figure}

\clearpage
\FloatBarrier
\subsection*{The causal graph structure of the Perturb-seq data}\label{secA3}

\begin{figure}[H]
  \centering
  \includegraphics[width=\linewidth]{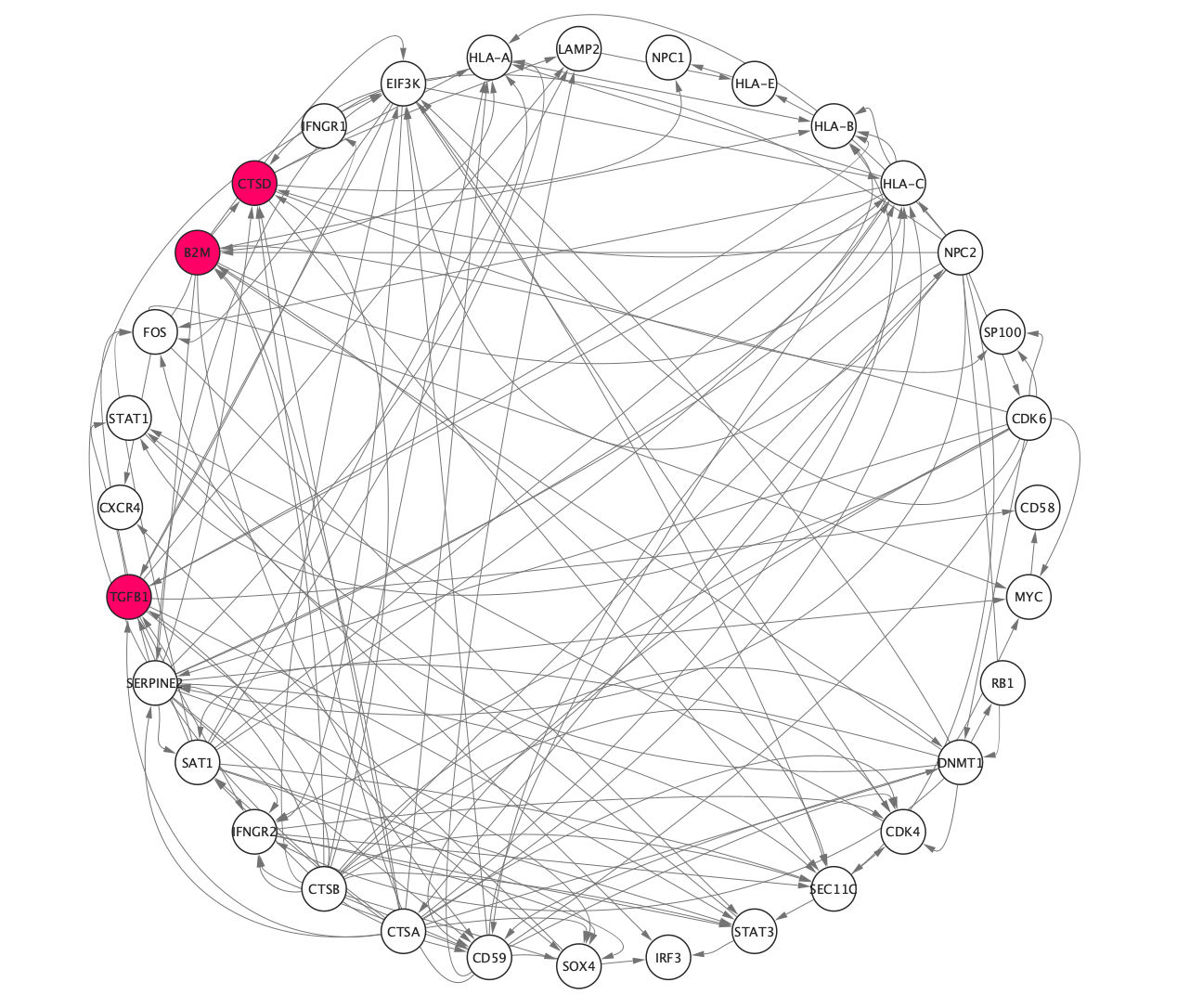}
  \caption{Directed acyclic graph (DAG) structure learnt over 36 genes using the greedy sparsest permutation (GSP) algorithm with known module-to-program regulatory relationships as prior information. Genes that are ground truth perturbation targets of the 3 case studies are highlighted.}
  \label{dag36}
\end{figure}

\clearpage
\subsection*{Supplementary results of COAST on the single-cell RNA-seq data}\label{secA4}

\begin{figure}[H]
  \centering
  \includegraphics[width=\linewidth]{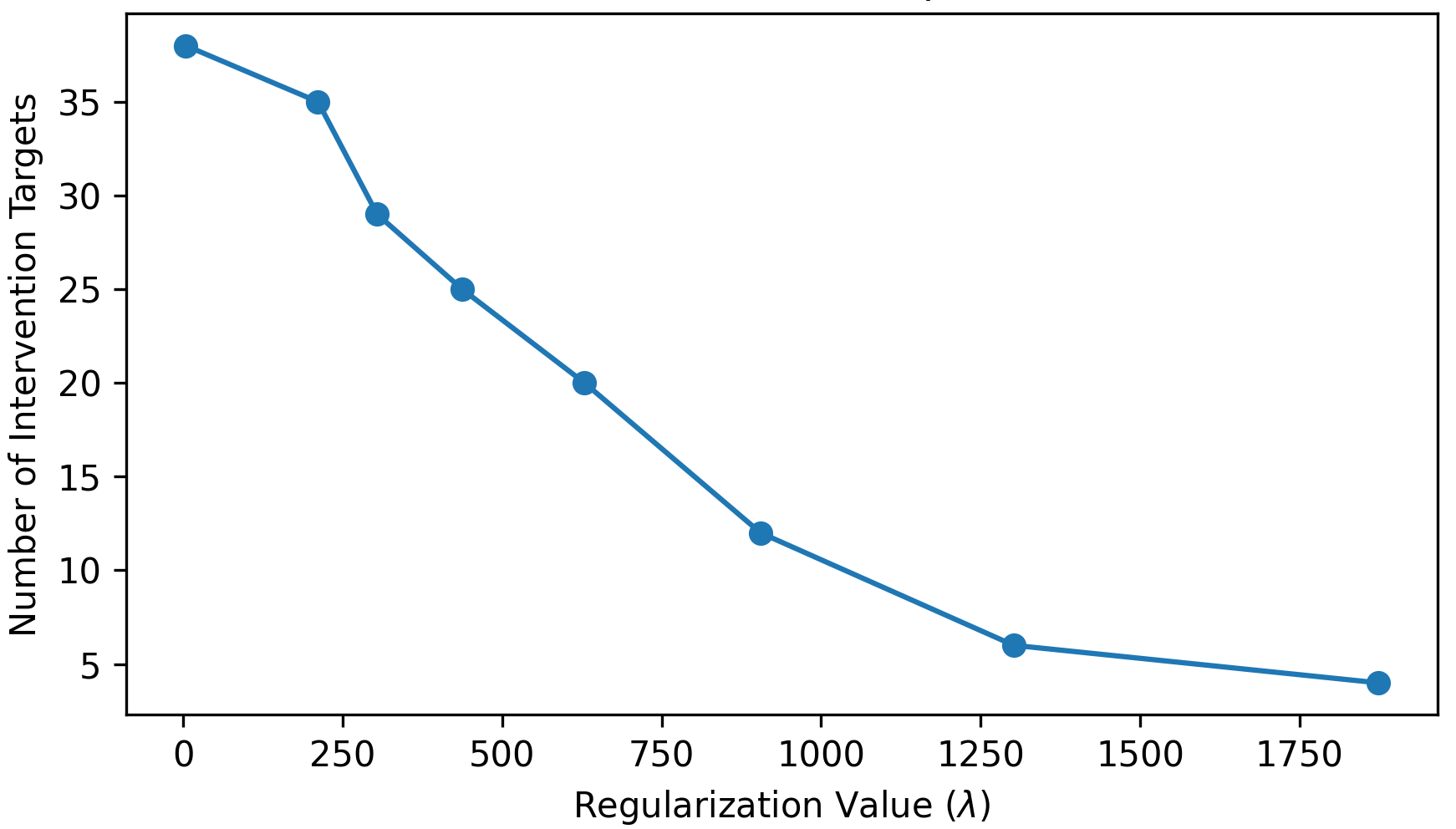}
  \caption{Results of COAST on the scRNA-seq dataset across a range of regularization strengths. As regularization increases, the number of inferred intervention targets progressively decreases from 38 (full candidate target set) to 4.}
  \label{results_reg_chen}
\end{figure}

Figure \ref{results_reg_chen} shows the results of COAST using the regularization approach on the scRNA-seq datasets in \cite{chen2017}. With a series of regularization values, the numbers of intervention targets of the identified solutions are 38, 35, 29, 24, 20, 12, 6, 4.

\FloatBarrier

\end{document}